%% file: main.tex
\documentclass{article}

\usepackage[dblblindworkshop, final]{neurips_2026}

\usepackage[utf8]{inputenc} 
\usepackage[T1]{fontenc}    
\usepackage{hyperref}       
\usepackage{url}            
\usepackage{booktabs}       
\usepackage{amsfonts}       
\usepackage{nicefrac}       
\usepackage{microtype}      
\usepackage{xcolor}         
\usepackage{graphicx}
\usepackage{amsmath}
\usepackage{booktabs}   
\usepackage{multirow}   
\usepackage[table]{xcolor} 
\usepackage{tabularx}
\usepackage{mdframed}
\usepackage{xcolor}
\usepackage{verbatim}
\usepackage{caption}
\definecolor{promptbg}{HTML}{F9F9F9}
\definecolor{promptframe}{HTML}{DDDDDD}
\usepackage{tcolorbox}
\tcbuselibrary{breakable}
\usepackage{listings}
\usepackage{graphicx}     
\usepackage{subcaption}   
\usepackage{natbib}       
\usepackage{booktabs}
\usepackage{soul}

\title{MAPLE: A Multimodal Mobile Agent with Playbook-Learned Experience}
\workshoptitle{VLM4RWD}

\author{%
  Jun Wang$^{1,2}$ 
  \quad Aurora Wei$^{1}$
  \quad Lucas Spangher$^{1}$
  \quad Ming Zhang$^{1}$ \\[0.3em]
  {\bfseries Eugene Ie$^{1}$ \qquad Yiannis Kantaros$^{2}$} \\[0.5em]
  \texttt{\{junw,ioannisk\}@wustl.edu} \\
  \texttt{\{aurorawei,spangher,mingzhang,eugeneie\}@google.com} \\[0.4em]
  $^{1}$Google
  \qquad $^{2}$Washington University in St. Louis
}

\begin{document}

\maketitle

\begin{abstract}
Autonomous agents powered by vision-language models (VLMs) show great potential in navigating mobile graphical user interfaces (GUIs) to complete complex tasks. However, existing paradigms face fundamental scalability bottlenecks: fine-tuning requires massive high-fidelity datasets or prohibitive training time and compute, while gradient-free approaches rely on custom static heuristics or costly trial-and-error trajectory exploration where feedback is deferred until the end of execution.
%
%
To address these limitations, we introduce MAPLE (\textit{Multimodal Agent with Playbook-Learned Experience}), a novel gradient-free multi-agent framework that enables dynamic, step-wise experience sharing and execution-time prompt evolution. Unlike traditional action-reflection architectures that depend on costly, post-hoc self-reflection, MAPLE enables agents to cross-reference and distribute runtime insights in real time at each operational step. This dynamic exchange eliminates blind trial-and-error looping and steers agents along direct, grounded interaction paths. Extensive evaluations on the AndroidWorld and MobileWorld benchmarks demonstrate that MAPLE achieves state-of-the-art performance on complex, long-horizon mobile tasks, while proving scalable and adaptable across novel environments.

\end{abstract}

\input{intro}

\input{methods}

\input{experiments}

\section{Conclusion and Future Work}
\label{sec:conclusion}

In this work, we presented MAPLE (Multimodal Agent with Playbook-Learned Experience), a gradient-free architecture designed for cold-start, continual experience accumulation across mobile GUIs. By decoupling intra-episodic state tracking via bounded Working Memory ($m_t$) from cross-episodic distillation via a persistent Playbook ($\mathcal{D}$), MAPLE transitions the mobile agent paradigm from brittle procedural memorization to the transfer of grounded, declarative UI mechanics. Comprehensive evaluations on AndroidWorld and MobileWorld demonstrate that MAPLE achieves leading training-free success rates---including top performance on user-interactive workflows---while guiding agents along more direct and purposeful execution trajectories. Crucially, our scaling analysis reveals that MAPLE's structured scaffolding provides disproportionate gains for lightweight and open-weight VLMs, preventing compounding failures under interactive ambiguity. Future work will explore adaptive scaffolding mechanisms to alleviate the Scaffolding Paradox on frontier models, as well as extending MAPLE's grounded experience-accumulation framework to broader digital control domains, including web browsing and cross-platform desktop automation.

\clearpage
\bibliographystyle{plain}
\bibliography{references}


\clearpage
\appendix

\input{appendix}



\end{document}

%% file: intro.tex

\section{Introduction}
\label{sec:intro}

Despite significant advancements in intuitive interface design, mobile devices remain challenging or inaccessible to large segments of the population, particularly the elderly and individuals with visual impairments. With the rapid evolution of Vision-Language Models (VLMs)~\cite{bordes2024introduction}, GUI-based autonomous agents have emerged as a transformative paradigm to bridge this accessibility gap and democratize access to essential modern technology. By empowering users to delegate complex, multi-application tasks via natural language, these agents hold immense potential to transform digital autonomy. Consequently, substantial research has focused on deploying foundation models as interactive reasoning engines across mobile platforms~\cite{wen2024autodroid,deng2023mind2web,wang2024mobile,hong2024cogagent,jiang2026openphone}, accompanied by comprehensive benchmarks~\cite{rawles2025androidworld,kong2026mobileworld,trivedi2024appworld} designed to evaluate the frontiers of agentic capabilities.

However, developing autonomous agents for mobile UI navigation requires balancing generalizability, online interaction overhead, and offline data curation costs. Existing paradigms struggle to strike this balance: gradient-based reinforcement learning requires prohibitively expensive environment rollouts, while zero-shot inference frameworks remain handicapped by static prompts or brittle, text-bound memory structures that fail to capture the visual nuances of dynamic mobile interfaces. In opaque, highly dynamic mobile GUIs, purely memorizing procedural step sequences is insufficient; an agent must dynamically reverse-engineer interface mechanics through visual transitions.

To bridge this fundamental gap, we introduce the \textit{Multimodal Agent with Playbook-Learned Experience} (\textbf{MAPLE}). Designed as a cold-start, continual learning system, MAPLE adapts the principles of step-level reflection and cross-episodic playbooks specifically for the multimodal mobile domain. Rather than relying on massive pre-curated datasets or text-only procedural memory, MAPLE leverages explicit visual state transitions (e.g., before-and-after screenshots) to autonomously distill declarative UI rules on the fly. By decoupling intra-episodic state tracking from cross-episodic knowledge accumulation and grounding retrieved priors in the active visual interface, MAPLE achieves robust, cold-start generalization across unseen mobile applications.

    
    
In summary, the primary contributions of this work are three-fold:
\begin{itemize}
    \item \textbf{Dual-Memory Cognitive Architecture:} We introduce MAPLE, a cold-start, gradient-free mobile agent architecture built around two complementary memory streams: (1) a \textit{bounded Working Memory} instantiated as a dynamic key-value store that tracks transient, within-episode task state (e.g., extracted data and sub-goal checklists), and (2) a \textit{persistent Playbook} that accumulates reusable, declarative UI mechanics across episodes. Coupled within a factorized step-level reasoning loop, these two memories coordinate short-term task execution with long-term interface priors without offline datasets or task-specific heuristics.
    
    \item \textbf{Self-Supervised Hindsight Distillation \& Grounded Retrieval:} We design a closed-loop mechanism to populate and query the Playbook entirely on the fly. By comparing post-action screenshot transitions against predicted visual outcomes, MAPLE autonomously distills and deduplicates universal action-effect facts, then retrieves them via intent masking and greedy visual anchor matching so injected guidance stays anchored to the active screen layout.
    
    \item \textbf{Extensive Evaluation \& State-of-the-Art Performance:} We evaluate MAPLE on two open-domain mobile benchmarks (\textit{AndroidWorld}~\cite{rawles2025androidworld} and \textit{MobileWorld}~\cite{kong2026mobileworld}) across six Vision-Language Models spanning open-weight to frontier scales. Powered by Gemini-3-Flash, MAPLE reaches 75.9\% Success Rate on AndroidWorld and 60.25\% on MobileWorld (including a record 68.18\% on interactive tasks), demonstrating superior navigational directness and strong cross-model generalization.
\end{itemize}

\section{Related Work}
\label{sec:related_work}

Developing autonomous agents for mobile UI navigation requires balancing policy generalizability, online interaction overhead, and offline curation costs. Existing literature generally bifurcates into gradient-based policy optimization and inference-only framework orchestration.

\subsection{Gradient-Based Policy Optimization in GUIs}
The first lineage leverages gradient-based reinforcement learning (RL) to cast GUI task execution as a policy optimization problem. Frameworks such as K$^2$-Agent~\cite{wu2026k}, MobileRL~\cite{xu2025mobilerl}, SoLS~\cite{papoudakis2026succeed}, and the GUI-R1 family~\cite{luo2025gui} explicitly fine-tune foundation models via algorithms such as Group Relative Policy Optimization (GRPO)~\cite{shao2024deepseekmath} or off-policy actor-critic methods with experience replay. While these approaches can discover robust interface policies, their scalability remains bottlenecked by severe sample inefficiency, as policy convergence demands extensive environmental interaction. On mobile platforms, executing rollouts over high-dimensional GUI state spaces inside live Android emulators incurs steep computational latency and infrastructure overhead, making large-scale, emulator-in-the-loop RL post-training prohibitively expensive.

\subsection{Gradient-Free \& Static Inference-Based Agents}
To bypass gradient optimization, a second lineage focuses on gradient-free, inference-based architectures leveraging zero-shot foundation models. While computationally lightweight, their adaptability to unseen domains is often hindered by reliance on static priors. Early approaches depended on handcrafted prompt templates or pre-collected UI exploration graphs (e.g., AppAgent~\cite{zhang2025appagent}, AutoDroid~\cite{wen2024autodroid}). Recent efforts like GEPA~\cite{agrawal2025gepa} automate prompt tuning via offline genetic optimization. However, these strategies rely on mini-batch sampling, parallel environment cloning, and static state assumptions—conditions that rarely hold across dynamic, stateful mobile applications.

\subsection{Dynamic Memory \& Continual Learning in Agents}
To eliminate offline curation, recent works explore continuous learning and dynamic memory architectures that update agent logic on the fly. For instance, Agentic Context Engineering (ACE)~\cite{zhang2025agenticcontextengineeringevolving} constructs a persistent, cross-session playbook to mitigate context collapse, while SCOPE~\cite{pei2025scope} introduces dual-stream tactical and strategic updates to recover from mid-task failures. 

Despite these advances, adapting text-centric memory frameworks like ACE and SCOPE to mobile GUIs reveals a fundamental limitation: they accumulate \textit{procedural task strategies} (memorizing specific action sequences) suited primarily for transparent, text-based environments. In opaque, highly dynamic visual GUIs, procedural memory generalizes poorly across UI updates and layout shifts. In contrast, MAPLE transitions from procedural memorization to \textit{causal rule distillation}. By evaluating multimodal state transitions (e.g., screenshot diffs), MAPLE autonomously extracts declarative facts about UI mechanics to construct a localized, visually-grounded world model.

%% file: methods.tex
\begin{figure}[t]
    \centering
    \includegraphics[width=0.9\textwidth]{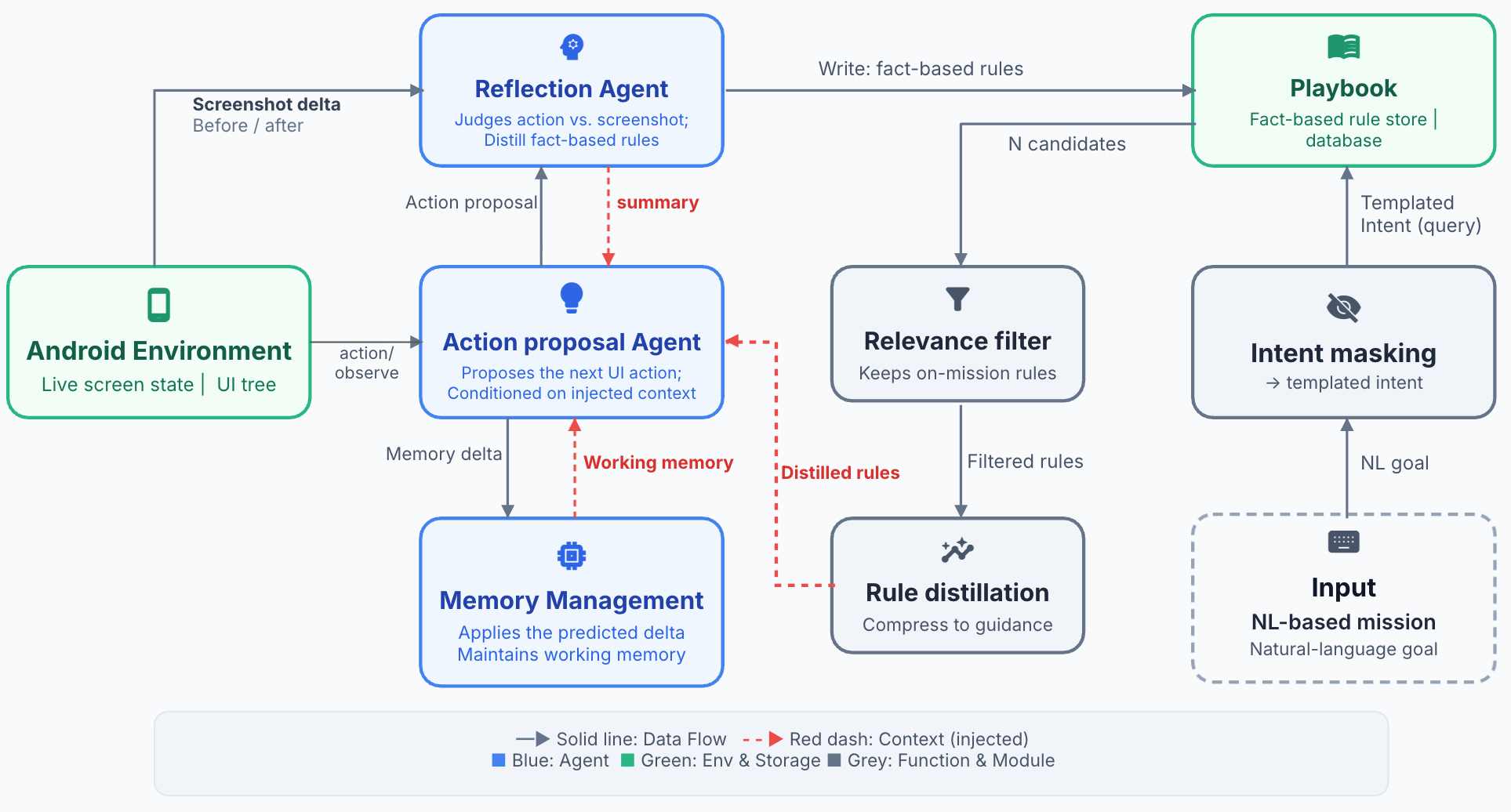}
    \caption{\textbf{Overview of the MAPLE framework.} MAPLE integrates intra-episodic UI execution with a cross-episodic learning loop. Given a user goal, the agent retrieves visually grounded priors from a persistent \textit{Playbook} and conditions on short-term \textit{Working Memory} to execute actions. Post-execution reflection updates local state tracking while distilling generalizable rules back into the Playbook for continuous knowledge accumulation.}
    \label{fig:maple_workflow}
\end{figure}

\section{Multimodal Agent with Playbook-Learned Experience (MAPLE)}
\label{sec:maple_architecture}

Figure~\ref{fig:maple_workflow} illustrates the overall execution and learning loop of \textbf{MAPLE}. We first formalize mobile GUI navigation as a dual-memory POMDP (Section~\ref{sec: problem_formulation}), then detail intra-episodic state tracking and factorized action selection (Sections~\ref{sec:working_memory}--\ref{sec:policy_factorization}), followed by cross-episodic Playbook distillation and visually grounded retrieval (Sections~\ref{sec:playbook_distillation}--\ref{sec:playbook_retrieval}).

\subsection{Problem Formulation}
\label{sec: problem_formulation}
We formalize mobile GUI navigation as a goal-conditioned Partially Observable Markov Decision Process (POMDP), defined by the tuple $\mathcal{M} = \langle \mathcal{S}, \mathcal{A}, \mathcal{T}, \Omega, \mathcal{O}, \mathcal{G}, \mathcal{R} \rangle$, where $\mathcal{S}$ denotes the unobservable system state space (including internal app variables and off-screen widgets), $\mathcal{T}$ the state transition dynamics, $\Omega$ the observation space, and $\mathcal{G}$ the space of natural-language user goals.
At step $t$, given latent state $s_t \in \mathcal{S}$, the observation process $\mathcal{O}$ coupled with Set-of-Marks (SoM) grounding converts the raw screen and view hierarchy into a multimodal observation $o_t = (\tilde{I}_t, E_t) \in \Omega$, pairing the tag-annotated screenshot $\tilde{I}_t$ with the parsed set of interactable UI elements $E_t$. Accordingly, the action space $\mathcal{A}$ comprises parameterized UI and dialogue operations (e.g., tap, swipe, text input, user query) indexed directly over $e \in E_t$, avoiding continuous coordinate regression. Because $s_t$ is latent, a standard agent conditions its policy $a_t \sim \pi(a_t \mid h_t, g)$ on the sampled goal $g \in \mathcal{G}$ and interaction history $h_t = (o_0, a_0, \dots, o_t)$ to generate an episode trajectory $\tau = (s_0, o_0, a_0, \dots, s_T)$ toward a terminal state $s_T$ satisfying $g$.

While task completion is evaluated only at episode termination via a sparse binary reward $\mathcal{R}(s_T, g) = \mathbb{I}(s_T \models g) \in \{0, 1\}$, a training-free VLM agent cannot optimize $\pi$ directly over full trajectories; instead, global task success depends on sequentially selecting well-grounded actions at each step $t$. Conditioning step-wise inference on the raw, append-only history $h_t$, however, accumulates irrelevant visual clutter and compounding false premises---early misinterpretations persist in context and derail downstream decisions---while purely myopic selection from $o_t$ alone gets trapped in unfamiliar UI layouts. To make step-wise action selection reliable over horizon $T$ without parameter updates, MAPLE reshapes the conditioning context through two complementary memory streams: (1) a persistent, cross-episodic \textit{Playbook} $\mathcal{D}_\ell$ (initialized cold-start at $\mathcal{D}_0 = \emptyset$), updated after each episode trajectory $\tau_\ell$ with reusable UI transition rules distilled via self-supervised hindsight reflection; and (2) a bounded, intra-episodic \textit{Working Memory} $m_t = \Phi(m_{t-1}, \Delta m_{t-1})$ (reset each episode at $m_0 = \emptyset$), where a deterministic parser $\Phi$ merges autoregressive key-value updates $\Delta m_{t-1}$ (insertions, modifications, or deletions) emitted at the previous step. At step $t$, combining $m_t$ with screen-grounded priors $\mathcal{P}_t$ retrieved from $\mathcal{D}$ forms a bounded context $C_t = \langle g, o_t, m_t, \mathcal{P}_t \rangle$, yielding the active policy $\pi(a_t \mid h_t, g) \approx \pi(a_t \mid C_t)$.

\subsection{Bounded Working Memory}
\label{sec:working_memory}
Rather than appending uncurated observation history, MAPLE instantiates the Working Memory ($m_t$) as a bounded, dynamic dictionary of semantic key-value pairs (e.g., \texttt{\{"clipboard": "code\_1", "item\_2\_status": "checked"\}}). Crucially, rather than relying on predefined schemas, both the keys and values (e.g., active sub-goals, runtime variables, and checklist states) are generated on-the-fly: at step $t$, the policy autoregressively predicts a schema update $\Delta m_t$, which the deterministic parser $\Phi$ merges into $m_t$.
This design replaces linearly growing raw screenshot histories with a compact, editable key-value state representation that remains interpretable and dynamically adapts to the active UI task.


\subsection{Autoregressive Policy Factorization}
\label{sec:policy_factorization}
Directly predicting actions via monolithic probability distributions $P(a_t \mid C_t)$ frequently leads to spatial ungrounding and temporal myopia. MAPLE factorizes the policy output into an autoregressive sequence of cognitive artifacts: 
\begin{equation}
\begin{split}
    P(a_t, \hat{o}_{t+1}, z_t, \Delta m_t \mid C_t) &= P(\Delta m_t \mid C_t) \cdot P(z_t \mid C_t, \Delta m_t) \\
    &\quad \cdot P(\hat{o}_{t+1} \mid C_t, \Delta m_t, z_t) \cdot P(a_t \mid C_t, \Delta m_t, z_t, \hat{o}_{t+1})
\end{split}
\end{equation}
Under this factorized bottleneck, the VLM generates a rigorous reasoning chain: (1) \textbf{Memory Modulation ($\Delta m_t$):} The policy emits memory deltas to iteratively update the working memory state $m_{t+1} = \Phi(m_t, \Delta m_t)$, appending, updating, or deleting semantic trackers; (2) \textbf{Chain-of-Thought Rationale ($z_t$):} Conditioned on the updated state, the policy generates a localized rationale aligning immediate visual evidence with the retrieved Playbook priors $\mathcal{P}_t$; (3) \textbf{Forward Dynamics Prediction ($\hat{o}_{t+1}$):} Crucially, the policy must explicitly hypothesize the \textit{semantic} visual consequence of its intended action prior to execution (e.g., predicting the appearance of a dropdown menu); and (4) \textbf{Action Execution ($a_t$):} Only after establishing this reasoning sequence does the policy emit the final parameterized interaction $a_t \in \mathcal{A}$ mapped to a discrete SoM element (e.g., \texttt{tap(element\_14)}). By architecturally requiring forward dynamics prediction ($\hat{o}_{t+1}$) and explicit state tracking ($\Delta m_t$) before mechanical execution, MAPLE ensures that physical interactions are strictly grounded in both episodic intent and historical knowledge, {reducing the risk of blind, repetitive looping behaviors commonly observed in unbottlenecked vision-language policies.}

\subsection{Self-Supervised Hindsight Distillation \& Playbook Consolidation}
\label{sec:playbook_distillation}

Following action execution $a_t$ and environment stabilization to observation $o_{t+1}$, MAPLE enters the hindsight reflection phase. While intra-episodic Working Memory ($m_t$) tracks transient task variables, cross-episodic learning requires evaluating observed state transitions to extract generalizable knowledge. MAPLE achieves this through a self-supervised reflection mechanism that distills transition dynamics into a dual-output hindsight artifact, consolidating universal interface mechanics into a persistent, non-parametric database termed the \textit{Playbook}.

\paragraph{Playbook Database Formalization.}
We define the Playbook $\mathcal{D} = \{d_1, d_2, \dots, d_N\}$ as a vector-indexed knowledge repository of verified UI mechanics, where each entry is a 5-tuple $d_i = (k_i, A_i, f_i, \mathbf{v}_i, c_i)$ comprising a generalized task intent template $k_i$, structural layout anchors $A_i$ (e.g., textual and state element descriptors) for spatial grounding, a natural-language declarative fact $f_i$ with dense semantic embedding $\mathbf{v}_i$, and an empirical confidence counter $c_i \in \mathbb{N}^+$ tracking rule verification and recurrence across episodes.

\paragraph{Dual-Output Hindsight Distillation.}
To evaluate environmental feedback autonomously, a reflection module compares the empirical transition tuple $(o_t, a_t, o_{t+1})$ against the agent's prior visual expectation $\hat{o}_{t+1}$.
From this comparison, it distills two complementary outputs $(\text{sum}_t, f_{\text{new}})$:
\begin{enumerate}
    \item \textbf{Localized Episodic Summary ($\text{sum}_t$):} A concise temporal description of the immediate transition (e.g., \textit{``Tapped `Filter', expanding the options dropdown''}). The summary $\text{sum}_t$ updates the intra-episodic trajectory history, providing immediate sequential context for subsequent steps in the active mission.
    \item \textbf{Universal Declarative Fact ($f_{\text{new}}$):} An abstracted, third-person action-effect rule defining an invariant UI property (e.g., \textit{``The vertical ellipsis icon expands the app configuration menu''}). By omitting task-specific instance details, $f_{\text{new}}$ functions as a modular world-model heuristic designed for zero-shot transfer across tasks.
\end{enumerate}

\paragraph{Semantic Consolidation and Deduplication.}
Newly distilled facts $f_{\text{new}}$ are packaged into a candidate entry $d_{\text{new}} = \big(k_{\text{intent}}, A(o_t), f_{\text{new}}, \mathbf{v}_{\text{new}}, 1\big)$ using the masked goal intent $k_{\text{intent}} = \psi(g)$, pre-action screen anchors $A(o_t)$, and dense fact embedding $\mathbf{v}_{\text{new}}$, then routed through a multi-stage consolidation pipeline before commitment to $\mathcal{D}$. Restricting to existing entries with matching intent $\{d_i \in \mathcal{D} \mid k_i = k_{\text{intent}}\}$, the system finds the nearest neighbor $i^* = \arg\max_{i:\, k_i = k_{\text{intent}}} \cos(\mathbf{v}_{\text{new}}, \mathbf{v}_i)$ with peak similarity $S_{\text{cos}}^* = \cos(\mathbf{v}_{\text{new}}, \mathbf{v}_{i^*})$ (setting $S_{\text{cos}}^* = 0$ if no entry yet shares $k_{\text{intent}}$). Integration into $\mathcal{D}$ is governed by dual similarity thresholds $\gamma_{\text{novel}}$ and $\gamma_{\text{dup}}$ ($0 < \gamma_{\text{novel}} < \gamma_{\text{dup}} < 1$):
\begin{equation}
    \mathcal{D} \leftarrow 
    \begin{cases} 
        \text{Update: } c_{i^*} \leftarrow c_{i^*} + 1, & \text{if } S_{\text{cos}}^* \ge \gamma_{\text{dup}} \quad \text{(Redundant)} \\
        \text{Insert: } \mathcal{D} \cup \{d_{\text{new}}\}, & \text{if } S_{\text{cos}}^* < \gamma_{\text{novel}} \quad \text{(Novel)} \\
        \text{Resolve}(\mathcal{D}, d_{\text{new}}, i^*), & \text{otherwise } (\gamma_{\text{novel}} \le S_{\text{cos}}^* < \gamma_{\text{dup}}),
    \end{cases}
\end{equation}
where $\text{Resolve}(\cdot)$ triggers an auxiliary semantic verification check. Specifically, the model evaluates whether $f_{\text{new}}$ and $f_{i^*}$ convey functionally identical UI mechanics. If deemed equivalent, the existing rule's confidence counter is incremented ($c_{i^*} \leftarrow c_{i^*} + 1$); otherwise, $d_{\text{new}}$ is inserted into $\mathcal{D}$ as a distinct rule. Through this closed-loop reflection and consolidation process, MAPLE maintains immediate episodic awareness while progressively accumulating a clean, deduplicated repository of reusable UI knowledge.

\subsection{Intent-Masked and Visually-Grounded Retrieval}
\label{sec:playbook_retrieval}

Prior to action execution at each decision step $t$, MAPLE queries the Playbook $\mathcal{D}$ using the goal $g$ and current observation $o_t$ to construct the heuristic prior $\mathcal{P}_t$. Naive prior retrieval using raw text or unconstrained features causes context saturation and hallucinations. To ensure retrieved priors are physically actionable on the active screen, MAPLE filters the Playbook $\mathcal{D}$ through a sequential two-stage retrieval pipeline:

\textbf{1. Semantic Intent Masking:} Raw user objectives $g$ often contain volatile, instance-specific entities (e.g., names, dates) that hinder direct matching. To enable cross-task generalization, MAPLE applies an abstraction function $k_{\text{intent}} = \psi(g)$ that maps specific inputs to canonical placeholders (e.g., \textit{``Send `123' to John''} becomes \textit{``Send \textlangle MESSAGE\textrangle\ to \textlangle RECIPIENT\textrangle''}). Using $k_{\text{intent}}$ as a high-level query key, we retrieve a task-relevant candidate subset $\mathcal{D}_{\text{cand}} = \{d_i \in \mathcal{D} \mid k_i = k_{\text{intent}}\}$.

\textbf{2. Greedy Visual Anchor Matching:} To guarantee that candidates in $\mathcal{D}_{\text{cand}}$ are physically applicable to the current interface, we evaluate their structural congruence with the active screen. We extract a set of active layout anchors $A(o_t) = \{u_1, \dots, u_M\}$ from the observation $o_t$, where each anchor represents a visible UI element $u_j = \langle \text{class}, \text{text}, \text{state} \rangle$. For each candidate $d_i \in \mathcal{D}_{\text{cand}}$ containing historically recorded anchors $A_i$, we compute a directed visual alignment score using a greedy maximum-similarity mapping: $S_{\text{anchor}}(A(o_t), A_i) = \frac{1}{|A_i|} \sum_{u \in A_i} \max_{u' \in A(o_t)} \text{sim}(u, u')$, where $\text{sim}(\cdot, \cdot) \in [0, 1]$ is a bounded semantic similarity score over anchor attributes. Candidates are pruned unless they satisfy a global visual congruence threshold: $S_{\text{anchor}} \ge \gamma_{\text{anchor}}$.

\textbf{Dynamic Prior Distillation.}
If no candidate satisfies $\gamma_{\text{anchor}}$ (e.g., at cold start), we set $\mathcal{P}_t = \emptyset$. When multiple candidates qualify, injecting raw, disjoint rules directly into the prompt can cause context redundancy or conflicting guidance. To resolve this, MAPLE ranks the qualified candidates by their visual score $S_{\text{anchor}}$ and historical confidence $c_i$, isolating the top-$K$ subset of declarative facts $\mathcal{F}_{\text{top-}K}$.
An auxiliary model then synthesizes the top-$K$ facts $\mathcal{F}_{\text{top-}K}$, conditioned on $g$ and active anchors $A(o_t)$, into a single cohesive heuristic prior $\mathcal{P}_t$. This distilled prior is appended directly to the action formulation context tuple $C_t = \langle g, o_t, m_t, \mathcal{P}_t \rangle$. By coupling semantic abstraction with real-time layout verification, this pipeline ensures that injected playbook guidance remains grounded in the active visual observation.

%% file: experiments.tex
\section{Experiments}
\label{sec:experiments}


\subsection{Experimental Setup}
\label{sec:experiment_setup}

\paragraph{Evaluation Benchmarks \& Setup.}
We evaluate MAPLE on two standardized mobile environments: (1) \textbf{AndroidWorld}~\cite{rawles2025androidworld}, comprising 116 programmatic tasks across 20 applications with dynamic task parameterization; and (2) \textbf{MobileWorld}~\cite{kong2026mobileworld}, a harder suite testing multi-app orchestration (averaging 27.8 steps vs. 14.3 in AndroidWorld) and interactive ambiguity. We evaluate on MobileWorld's 161 core tasks across its GUI-Only (117 tasks) and Agent-User Interaction (44 tasks) splits. Across both benchmarks, agents issue parameterized UI actions over Set-of-Marks grounded elements. An episode terminates upon the agent signaling completion or exhausting the step budget $T$, with task success $\mathbb{I}(s_T \models g)$ verified deterministically by native benchmark evaluators auditing ground-truth system states (e.g., querying SQLite databases).

\paragraph{Metrics \& Protocol.}
An evaluation episode terminates upon the agent issuing a completion action or exhausting the pre-allocated step budget $T$. Following standard benchmark conventions, we evaluate performance across two primary metrics:
\begin{itemize}
    \item \textbf{Task Success Rate (SR, \%):} The proportion of tasks deterministically completed within $t \le T$ steps, as verified by native benchmark environment evaluators.
    \item \textbf{Average Trajectory Length (Steps):} The average number of physical UI interactions required for task completion, quantifying navigational directness and the agent's ability to reach target states without trial-and-error looping.
\end{itemize}
For cold-start evaluation, the Playbook starts completely empty ($\mathcal{D}_0 = \emptyset$) at the beginning of each benchmark suite, accumulates distilled rules sequentially across a single pass in the benchmark's default task order (while Working Memory resets every episode, $m_0 = \emptyset$), and is wiped cleanly between model backbones and ablation runs to prevent cross-run leakage. Complete implementation details—including VLM hyper-parameters, exact step budgets, and prompt templates—are provided in Appendix~\ref{subsec:hyperparameters_run}.

\paragraph{Baselines.}
We benchmark MAPLE against two agent categories: 
\textbf{(1) Adapted M3A (Controlled Baseline):} We adapt the M3A framework~\cite{rawles2025androidworld}, which uses an \textit{Observe-Think-Act-Reflect} loop similar to \textit{Reflexion}~\cite{shinn2023reflexion} and \textit{Voyager}~\cite{wang2023voyager}. To evaluate true cross-app generalization, we remove handcrafted per-task guidelines while upgrading M3A with the same structured prompting (including the Discoverability Protocol; Appendix~\ref{appendix:discoverability_ablation}), Set-of-Marks grounding, and dialogue actions as MAPLE. Sharing the identical VLM backbone but lacking Working Memory ($m_t$), forward dynamics prediction ($\hat{o}_{t+1}$), and the Playbook ($\mathcal{D}$), this baseline directly isolates the gains of our architecture. 
\textbf{(2) Inference-Only SOTA:} We compare against published leaderboard baselines, including Agent-SAMA~\cite{guo2026agent} and MobileUse~\cite{li2025mobileuse}, which execute actions via static prompt contexts without dynamic memory or hindsight distillation.

\begin{table}[t]
\centering
\caption{\textbf{Main Results on Mobile Agent Benchmarks.} MAPLE demonstrates superior execution across AndroidWorld ($1\times$ step budget) and MobileWorld (161 tasks). In AndroidWorld, MAPLE significantly outperforms all strictly zero-shot/gradient-free academic baselines, performing on par with gradient-based RL frameworks. In the highly complex MobileWorld environment, despite utilizing a lightweight Flash backbone, MAPLE achieves \textbf{Rank \#3 Overall} and secures the \textbf{Rank \#1 SOTA on User-Interactive tasks} across the entire public leaderboard.}
\label{tab:main_results}
\resizebox{\textwidth}{!}{%
\begin{tabular}{@{} l l c c c c @{}}
\toprule
\multicolumn{6}{@{}l}{\textbf{(a) AndroidWorld Benchmark (116 Tasks)}} \\ \midrule
\multicolumn{2}{@{}l}{\textbf{Method / Agent}} & \multicolumn{2}{c}{\textbf{Backbone Model}} & \textbf{Paradigm} & \textbf{Success Rate (\%)} \\ \midrule
\textit{Unrestricted Leaderboard Frontiers$^\ddagger$} & & & & & \\
\multicolumn{2}{@{}l}{MobileUseAgent \cite{vephone_mobileuse_2024}} & \multicolumn{2}{c}{Seed1.8-GUI} & Unrestricted Agent & 97.4 \\
\multicolumn{2}{@{}l}{AutoDevice \cite{autodevice_androidworld_2026}} & \multicolumn{2}{c}{Gemini-3-Pro + Sonnet} & Unrestricted Agent & 94.8 \\ \midrule
\textit{Controlled Academic Baselines (Gradient-Free)} & & & & & \\
\multicolumn{2}{@{}l}{Stable Planner \cite{mo2025building}} & \multicolumn{2}{c}{Qwen2.5-VL-72B} & Zero-Shot & 63.8 \\
\multicolumn{2}{@{}l}{Agent-SAMA \cite{guo2026agent}} & \multicolumn{2}{c}{GPT-4o + Qwen-VL} & Zero-Shot & 63.7 \\
\multicolumn{2}{@{}l}{MobileUse \cite{li2025mobileuse}} & \multicolumn{2}{c}{Qwen2.5-VL-72B} & Zero-Shot & 62.9 \\
\multicolumn{2}{@{}l}{M3A Baseline (Adapted) \cite{rawles2025androidworld}} & \multicolumn{2}{c}{Gemini-3-Flash} & Zero-Shot & 71.6 \\
\rowcolor{gray!10} \multicolumn{2}{@{}l}{\textbf{MAPLE (Ours)}} & \multicolumn{2}{c}{\textbf{Gemini-3-Flash}} & Dual-Memory & \textbf{75.9} \\ \midrule
\textit{Controlled Academic Baselines (Gradient-Based)} & & & & & \\
\multicolumn{2}{@{}l}{K$^2$-Agent \cite{wu2026k}} & \multicolumn{2}{c}{Qwen2.5-VL-72B} & RL (Upper Bound) & 76.7 \\ \midrule
\multicolumn{6}{c}{} \\ [-0.5em] 
\toprule
\multicolumn{6}{@{}l}{\textbf{(b) MobileWorld Benchmark (161 Tasks: 117 GUI-Only + 44 User-Interactive)}} \\ \midrule
\textbf{Method / Agent} & \textbf{Backbone Model} & \textbf{Type} & \textbf{Overall (\%)} & \textbf{GUI-Only (\%)} & \textbf{User-Int. (\%)} \\ \midrule
\textit{Heavyweight \& Multi-Model Frontiers} & & & & & \\
MagicGUI-Agent \cite{tang2025magicgui} & GPT-5.4 + Seed-1.8 & Agentic & 66.5 & 66.7 & 65.9 \\
Seed-2.0-Pro \cite{seed2026seed2} & Seed-2.0-Pro & General & 62.7 & 63.2 & 61.4 \\
Claude-Opus-4.7 & Claude-Opus-4.7 & General & 57.1 & 56.4 & 59.1 \\
GPT-5 + UI-Ins-7B & GPT-5 + UI-Ins-7B & Agentic & 56.2 & 54.0 & 62.2 \\ \midrule
\textit{Controlled Comparison (Lightweight Backbone)} & & & & & \\
Gemini-3.1-Pro-Preview & Gemini-3.1-Pro-Preview & General & 48.4 & 58.1 & 22.7 \\
M3A Baseline (Adapted) & Gemini-3-Flash & Agentic & 53.42 & 53.85 & 52.27 \\
\rowcolor{gray!10} \textbf{MAPLE (Ours)} & Gemini-3-Flash & Dual-Memory & \textbf{60.25} & \textbf{57.26} & \textbf{68.18}$^\dagger$ \\ \midrule
\multicolumn{6}{@{}l}{\small $^\ddagger$ Unrestricted entries frequently utilize heavy manual heuristics or closed-source system hacks, serving as environment ceilings.} \\
\multicolumn{6}{@{}l}{\small $^\dagger$ Denotes \textbf{Rank \#1 SOTA} on the public MobileWorld User-Interactive benchmark.} \\
\multicolumn{6}{@{}l}{\small \textbf{Model Types:} \textit{Agentic}: planner-executor systems with a grounding module. \textit{General}: end-to-end LMMs without GUI fine-tuning.} \\
\bottomrule
\end{tabular}%
}
\end{table}

\subsection{Main Results}
\label{sec:results}

Our empirical evaluation investigates two central questions: (1) Does MAPLE's integrated dual-memory architecture outperform static zero-shot frameworks without relying on handcrafted, task-specific heuristics? and (2) How effectively does MAPLE generalize across long-horizon, multi-app, and interactive mobile workflows? Table~\ref{tab:main_results} summarizes end-to-end performance on \textit{AndroidWorld} and \textit{MobileWorld} using Gemini-3-Flash as the primary VLM backbone.

\paragraph{State-of-the-Art Generalization on AndroidWorld.}
As shown in Table~\ref{tab:main_results}(a), while unrestricted leaderboard entries (e.g., MobileUseAgent, AutoDevice) achieve high ceilings, they frequently rely on heavy environment-specific scaffolding, bespoke tool pipelines, or hardcoded per-app heuristic overrides. In contrast, operating strictly through standard Set-of-Marks observations without task- or app-specific guidance, MAPLE sets a strong bar for training-free mobile agents. MAPLE achieves a \textbf{75.9\% Success Rate (SR)}, significantly outperforming competitive gradient-free academic frameworks including MobileUse ($62.9\%$), Agent-SAMA ($63.7\%$), and Stable Planner ($63.8\%$). Remarkably, without requiring parameter updates or offline environment rollouts, MAPLE performs competitively with compute-intensive, gradient-fine-tuned reinforcement learning baselines such as $K^2$-Agent ($76.7\%$).

\paragraph{Strong Multi-App Performance on MobileWorld.}
MobileWorld introduces challenging long-horizon dependencies across multiple applications. As presented in Table~\ref{tab:main_results}(b), MAPLE achieves an \textbf{Overall SR of 60.25\%}, securing \textbf{Rank 3 Overall} on the public MobileWorld benchmark. Powered by a lightweight Gemini-3-Flash backbone, MAPLE's dual-memory architecture surpasses unaugmented frontier models including Gemini-3.1-Pro-Preview ($48.4\%$) and Claude-Opus-4.7 ($57.1\%$), as well as specialized agentic setups utilizing GPT-5 ($56.2\%$).

\begin{table}[htpb]
\centering
\caption{\textbf{Ablation Study on Model Generalization.} 
With identical Discoverability Protocol prompting (Appendix~\ref{appendix:discoverability_ablation}), MAPLE improves success rates over the M3A baseline across all four VLM backbones while reducing trajectory steps on three.}
\label{tab:model_ablation}
\resizebox{\textwidth}{!}{%
\begin{tabular}{@{} l l c c c c @{}}
\toprule
\textbf{Model} & \textbf{Method} & \textbf{Overall SR (\%)} & \textbf{GUI-Only (\%)} & \textbf{User-Int. (\%)} & \textbf{Trajectory Length} \\ \midrule
\rowcolor{gray!10} \textbf{Gemini-3-Flash} & \textbf{MAPLE (Ours)} & \textbf{60.25} & \textbf{57.26} & \textbf{68.18} & \textbf{24.14} \\
 & M3A Baseline & 53.42 & 53.85 & 52.27 & 25.65 \\ \midrule
\rowcolor{gray!10} \textbf{Gemini-3-Flash-Lite} & \textbf{MAPLE (Ours)} & \textbf{39.13} & \textbf{35.04} & \textbf{50.00} & \textbf{19.13} \\
 & M3A Baseline & 31.68 & 31.62 & 31.82 & 21.11 \\ \midrule
\rowcolor{gray!10} \textbf{Qwen3.6-35B} & \textbf{MAPLE (Ours)} & \textbf{27.95} & \textbf{28.21} & \textbf{27.27} & 26.17 \\
 & M3A Baseline & 25.47 & 27.35 & 20.45 & 25.23 \\ \midrule
\rowcolor{gray!10} \textbf{Gemma4-26B-A4B} & \textbf{MAPLE (Ours)} & \textbf{27.95} & \textbf{27.35} & \textbf{29.55} & \textbf{34.00} \\
 & M3A Baseline & 14.29 & 17.09 & 6.82 & 45.60 \\
\bottomrule
\end{tabular}%
}
\end{table}

\paragraph{Robustness under Interactive Ambiguity.}
By coordinating explicit state tracking ($m_t$), meta-cognitive ambiguity triggers, and grounded Playbook priors ($\mathcal{P}_t$), MAPLE prevents futile exploratory loops and preserves context bandwidth for resolving multi-turn dialogue ambiguity. Consequently, MAPLE achieves 68.18\% SR on User-Interactive tasks—a +15.91\% absolute improvement over the Adapted M3A baseline (52.27\%). This establishes the highest score recorded to date on this split, outperforming specialized frameworks such as MagicGUI-Agent (65.9\%).

\paragraph{Model Generalization Across Scales.}
To validate that MAPLE's gains are model-agnostic, we evaluate across proprietary (\texttt{Gemini-3-Flash}, \texttt{Gemini-3-Flash-Lite}) and open-weight (\texttt{Qwen3.6-35B}, \texttt{Gemma4-26B-A4B}) backbones against the controlled Adapted M3A baseline (Table~\ref{tab:model_ablation}) on MobileWorld. Across all model tiers, MAPLE consistently boosts task success while steering agents toward shorter, more direct interaction paths on three of the four backbones (and adding less than one step on \texttt{Qwen3.6-35B} despite completing substantially more multi-turn interactive tasks). The architectural gains are most pronounced on the ambiguous \textit{User-Interactive} split: MAPLE elevates \texttt{Gemini-3-Flash} to a peak \textbf{68.18\% SR} ($+15.91\%$), while on \texttt{Gemma4-26B-A4B}, it rescues the policy from catastrophic failure, quadrupling interactive success ($6.82\% \to 29.55\%$) and cutting trajectory steps by $25.4\%$ ($45.60 \to 34.00$). Consistent gains also hold across all six evaluated backbones on AndroidWorld (Appendix~\ref{subsec:androidworld_cross_arch}).

\paragraph{The Scaffolding Paradox and Limitations.}
When scaled to frontier models on MobileWorld, MAPLE plateaus or slightly regresses in overall success rate (matching \texttt{Claude-Opus-4.8} at $63.35\%$ and dipping from $62.73\%$ to $59.63\%$ on \texttt{Gemini-3.6-Flash}), even though it continues to boost both models on deterministic AndroidWorld tasks (Appendix~\ref{subsec:androidworld_cross_arch}). We attribute this saturation to the \textit{Scaffolding Paradox}: frontier models possess strong intrinsic reasoning and naturally excel at exploratory recovery (``smart wandering'') under UI ambiguity. Imposing structured Playbook priors acts as an overly rigid constraint, narrowing the model's search space and impeding opportunistic problem-solving. Crucially, MAPLE still shortens average trajectory length across both models ($23.24 \to 22.20$ and $21.70 \to 20.40$ steps), confirming that it enforces a direct navigation policy; however, for frontier models, aggressively pruning exploratory rollouts removes the exact trial-and-error mechanism required to resolve complex edge cases. Details in Appendix~\ref{subsec:scaffolding_paradox}.

%% file: appendix.tex
\begin{center}
    \Large \textbf{Appendix: Supplementary Material}
\end{center}
\vspace{0.5cm} 

\section*{Appendix Overview}
\label{appendix:overview}

To assist readers in navigating the supplementary material, this appendix provides comprehensive experimental details, extended analyses, and full system configurations for the MAPLE framework. The content is structurally partitioned into two main sections corresponding to our primary evaluation environments:

\begin{itemize}
    \item \textbf{Section \ref{appendix:mobile_world}: MobileWorld (Interactive \& Long-Horizon Evaluation).} This section details our primary testbed for handling ambiguity and cross-application workflows. 
    \begin{itemize}
        \item \textbf{Benchmark \& Setup:} Section \ref{appendix:mobileworld_details} defines the task taxonomy, while Section \ref{subsec:hyperparameters_run} lists all exact VLM hyperparameters.
        \item \textbf{Ablation Studies:} We provide in-depth analyses on the impact of user-interactive triggering (Section \ref{appendix:discoverability_ablation}), cross-architecture generalizability (Section \ref{subsec:mobileworld_cross_arch}), and the necessary VLM cognitive bandwidth/thinking budgets required for dynamic memory processing (Section \ref{subsec:ablation_thinking_budget}).
        \item \textbf{Limitations:} Section \ref{subsec:scaffolding_paradox} explores the ``scaffolding paradox,'' discussing where explicit memory constraints cause performance saturation in frontier models.
        \item \textbf{Prompts \& Visuals:} Section \ref{appendix:maple_prompt} provides the complete, unedited system prompts for all MAPLE modules, followed by a visual example of Set-of-Mark (SoM) grounding in Section \ref{subsec:mobileworld_observation}.
    \end{itemize}

    \item \textbf{Section \ref{appendix:android_world}: AndroidWorld (Deterministic \& Zero-Shot Evaluation).} This section outlines our validation on standard, heavily parameterized autonomous control tasks.
    \begin{itemize}
        \item \textbf{Ablations \& Benchmark Nuances:} Following benchmark details in Section \ref{subsec:androidworld_details}, we present further cross-architecture validations (Section \ref{subsec:androidworld_cross_arch}) and investigate how relaxing step budgets impacts autonomous error recovery (Section \ref{subsec:step_budget_ablation}).
        \item \textbf{Systemic Limitations Audit:} Section \ref{subsec:androidworld_limitations} provides a critical, programmatic audit of 13 ``hard-to-complete'' missions, detailing where benchmark rigidities and false negatives decouple from true agent reasoning. 
        \item \textbf{Configurations \& Visuals:} We detail our zero-shot hyperparameter and prompt adaptations in Sections \ref{subsec:androidworld_hyperparameters} and \ref{subsec:androidworld_prompts}, concluding with a step-by-step visual demonstration of a successful task trajectory in Section \ref{subsec:androidworld_trajectory}.
    \end{itemize}
\end{itemize}

\section{Mobile World}
\label{appendix:mobile_world}

\subsection{Benchmark Details}
\label{appendix:mobileworld_details}

To rigorously evaluate the generalization and multi-step reasoning capabilities of our proposed MAPLE agent, we utilize \textbf{MobileWorld} \cite{kong2026mobileworld}, a recently introduced, highly challenging benchmarking environment for autonomous mobile agents. While prior benchmarks like AndroidWorld have driven significant progress, their saturation (with state-of-the-art models exceeding 90\% success rates) and reliance on short-horizon tasks limit their utility for evaluating production-grade autonomy. MobileWorld directly addresses these shortcomings by emphasizing long-horizon workflows and introducing interactive, ambiguity-driven task paradigms. 

\paragraph{Task Taxonomy and Scope}
MobileWorld consists of tasks spanning 20 third-party applications, authentically replicating real-world usage patterns. For our evaluation, we focus on the core subset comprising 161 tasks, which is structurally partitioned into two primary categories:
\begin{itemize}
    \item \textbf{GUI-Only (117 tasks):} Traditional visual navigation and interface manipulation tasks, demanding precise visual grounding, memory retention, and spatial reasoning over extended trajectories.
    \item \textbf{Agent-User Interaction (44 tasks):} A novel paradigm featuring deliberate instructional ambiguity. In these tasks, critical information is intentionally omitted from the initial prompt, requiring the agent to proactively trigger an \texttt{ask\_user} action rather than failing silently or hallucinating.
\end{itemize}
\textit{(Note: We deliberately omit the 40 MCP-Augmented hybrid API tasks, as MAPLE is fundamentally designed to optimize cross-episodic visual GUI navigation and semantic intent masking.)}

MobileWorld introduces severe cross-application complexity: approximately 62.2\% of its tasks require orchestrating workflows across multiple apps. Consequently, the average trajectory length extends to 27.8 steps—nearly doubling AndroidWorld's average. This extended horizon strictly tests an agent's working memory capacity and resilience against cascading errors, directly motivating our introduction of the dynamic Playbook memory in MAPLE.

\paragraph{Deterministic Evaluation Framework}
MobileWorld operates within a containerized environment with full backend state access for deterministic evaluation. Task success is verified programmatically via backend database inspection (47.3\%) and local system storage analysis (36.8\%), completely bypassing fragile LLM-as-a-judge heuristics and ensuring absolute reliability in multi-stage transaction outcomes.

\subsection{Ablation Study on User-Interactive Triggering: Analyzing Benchmark Sensitivity and Meta-Cognitive Boundaries}
\label{appendix:discoverability_ablation}

A critical challenge in interactive mobile agent benchmarks such as MobileWorld is governing when an agent should act autonomously versus when it must solicit user clarification. In preliminary evaluations, we observed that agents confronted with ambiguous instructions or absent parameters frequently engaged in unproductive trial-and-error behaviors---such as searching on-device contacts, browsing local settings, or guessing missing values---rather than asking the user directly. 

To regulate this decision boundary, we introduced the \textbf{Discoverability Protocol} into the agent's system prompt (detailed in Appendix~\ref{appendix:maple_prompt}). The protocol explicitly forbids unprompted parameter hunting and enforces an immediate \texttt{ask\_user} action across three mutually exclusive conditions:
\begin{enumerate}
    \item \textbf{Subjective Preferences:} Vague or under-specified criteria (e.g., ``clean up useless files'') requiring explicit user definition before interaction.
    \item \textbf{Missing Hard Data:} Unspecified prerequisite credentials or identifiers (e.g., phone numbers, PINs) that must not be inferred via on-device search.
    \item \textbf{Execution Blocks:} Hard roadblocks such as authentication walls or dead-end queries where repeated autonomous retries induce loops.
\end{enumerate}
Furthermore, the protocol assigns this clarification check top priority in the agent's decision hierarchy, overriding standard navigational heuristics.

\begin{table}[h]
\centering
\small
\caption{Ablation on the Discoverability Protocol on MobileWorld User-Interactive tasks (44 tasks, \texttt{Gemini-3-Flash}). The last two rows match Table~1(b) and Table~2.}
\label{tab:protocol_ablation}
\begin{tabular}{llcc}
\toprule
\textbf{Method} & \textbf{Discoverability Protocol} & \textbf{User-Int. SR (\%)} & \textbf{Solved Tasks} \\
\midrule
M3A Baseline (Adapted) & Disabled & 31.82 & 14 / 44 \\
M3A Baseline (Adapted) & Enabled (Table~2) & 52.27 & 23 / 44 \\
\textbf{MAPLE (Ours)}  & Enabled (Table~2) & \textbf{68.18} & \textbf{30 / 44} \\
\bottomrule
\end{tabular}
\end{table}

The ablation results in Table~\ref{tab:protocol_ablation} reveal two primary findings:
\begin{enumerate}
    \item \textbf{Explicit clarification rules resolve ambiguity bottlenecks:} Without the three-rule Discoverability Protocol, the Adapted M3A baseline attains only a 31.82\% Success Rate (SR) on User-Interactive tasks. Introducing the protocol increases baseline SR to 52.27\% (+20.45 pp, as reported in Tables~1(b) and~2). This substantial gain confirms that explicit meta-cognitive triggers are indispensable for suppressing premature trial-and-error under under-specified instructions.
    \item \textbf{MAPLE delivers significant orthogonal improvements:} To maintain strict parity, the Discoverability Protocol is uniformly enabled across all baselines and MAPLE variants in our primary evaluations. With this baseline constraint held constant, MAPLE's holistic architecture---integrating bounded working memory ($m_t$), predictive dynamics ($\hat{o}_{t+1}$), and structured playbook memory ($\mathcal{D}$)---further boosts User-Interactive SR from 52.27\% to \textbf{68.18\%} (+15.91 pp).
\end{enumerate}
These results highlight the necessity of standardizing clarification directives when benchmarking interactive agents. Moreover, because both models share the identical protocol, MAPLE's downstream margins reflect architectural synergy across memory and world modeling rather than prompt-level clarification incentives alone.

\subsection{Ablation Study: Cross-Architecture Robustness and Generalization}
\label{subsec:mobileworld_cross_arch}

A critical benchmark for any agentic framework is demonstrating that its performance gains are intrinsic to MAPLE's coupled dual-memory and reflective reasoning architecture, rather than artifacts of a specific underlying Vision-Language Model's (VLM) capabilities. To rigorously validate the architectural generalizability of MAPLE, we conduct an ablation study across a diverse spectrum of model families, encompassing both proprietary (Gemini-3-Flash, Gemini-3-Flash-Lite) and open-weight (Qwen3.6-35B, Gemma4-26B-A4B) backbones. 

As detailed in Table~\ref{tab:model_ablation}, integrating MAPLE consistently yields a universal performance lift across all tested architectures, strongly corroborating its model-agnostic robustness. Importantly, these improvements scale effectively across different model capacities.

The framework's impact is most profound in the highly complex \textit{User Interactive} split, which heavily penalizes models lacking robust meta-cognitive policies. For the highly capable Gemini-3-Flash model, MAPLE pushes performance to state-of-the-art levels, yielding a 15.91\% absolute increase in success rate. Even more compelling is MAPLE's ability to ``rescue'' smaller open-weight models from compounding errors in ambiguous scenarios. When utilizing Gemma4-26B-A4B, the M3A baseline nearly collapses on interactive tasks (6.82\% SR); however, the introduction of MAPLE’s dynamic priors completely stabilizes the policy, resulting in a near-quadrupling of the success rate to 29.55\%.


Finally, MAPLE improves task success while finding shorter, more direct interaction paths on three of the four evaluated backbones. Rather than relying on brute-force exploration, the harness helps the agent reason more effectively about UI state transitions and avoid repetitive sub-goal loops. Most strikingly, Gemma4-26B-A4B nearly doubles its overall success rate while requiring 25.4\% fewer interaction steps (45.60 to 34.00). On Qwen3.6-35B, higher completion rates on complex User-Interactive workflows (20.45\% to 27.27\%) are achieved with a negligible shift in trajectory length (25.23 to 26.17 steps). Across backbones, these results indicate that MAPLE improves success not by taking more exploratory actions, but by steering the policy along cleaner, higher-precision decision paths.

\subsection{Ablation Study: Cognitive Bandwidth and Thinking Budget Prerequisites}
\label{subsec:ablation_thinking_budget}

A fundamental consideration when deploying advanced agentic frameworks is the computational overhead and cognitive bandwidth required to maintain and query dynamic memory structures. To investigate the dependency of MAPLE on the underlying Vision-Language Model's (VLM) internal reasoning capacity, we conducted an ablation study modulating the explicit ``thinking budget'' of the lightweight \textit{Gemini-3-Flash-Lite} model on the MobileWorld benchmark. 

The results, detailed in Table~\ref{tab:thinking_budget}, reveal a critical architectural prerequisite: MAPLE requires a sufficient internal thinking budget to effectively process its cross-episodic Playbook and multi-stage reflection pipeline. 

Under a \textbf{Minimal} thinking budget, MAPLE experiences a performance inversion, underperforming the M3A Baseline in Overall Success Rate (22.36\% vs. 29.19\%). This indicates that in severely compute-constrained, highly reactive settings, the additional context from intent-masked retrieval and visual grounding can overwhelm the model's limited reasoning capacity, leading to context collapse or superficial action formulation.

However, this initial limitation highlights MAPLE's distinct advantage in compute scaling. When the reasoning constraints are lifted (\textbf{High} thinking budget), the M3A Baseline exhibits marginal gains, plateauing from 29.19\% to 31.68\%. In stark contrast, MAPLE's architecture efficiently channels the increased cognitive bandwidth, surging to a 39.13\% Overall Success Rate. This scaling is most pronounced in the complex \textit{User-Interactive} split, where MAPLE jumps from 25.00\% to 50.00\%. 

Ultimately, this ablation demonstrates that while MAPLE introduces a baseline requirement for cognitive bandwidth, it fundamentally unlocks the VLM's ability to translate increased compute into superior long-horizon planning and ambiguity resolution—capabilities where standard zero-shot baselines simply plateau.

\begin{table}[htpb]
\centering
\small
\caption{Ablation on VLM Thinking Budget using Gemini-3-Flash-Lite on the MobileWorld benchmark. MAPLE strictly requires a High thinking budget to process its dynamic memory, effectively unlocking performance scaling where the baseline plateaus.}
\label{tab:thinking_budget}
\begin{tabular}{@{}llcccc@{}}
\toprule
\textbf{Budget} & \textbf{Method} & \textbf{Overall SR (\%)} & \textbf{GUI-Only SR (\%)} & \textbf{User-Int. SR (\%)} & \textbf{Avg. Steps} \\ \midrule
\rowcolor{gray!10} 
\textbf{Minimal} & MAPLE (Ours) & 22.36 & 21.37 & 25.00 & 15.24 \\
 & M3A Baseline & 29.19 & 29.91 & 27.27 & 17.05 \\ \midrule
\rowcolor{gray!10} 
\textbf{High} & MAPLE (Ours) & \textbf{39.13} & \textbf{35.04} & \textbf{50.00} & 19.13 \\
 & M3A Baseline & 31.68 & 31.62 & 31.82 & 21.11 \\ 
\bottomrule
\end{tabular}
\end{table}

\subsection{Limitations: The Scaffolding Paradox and Frontier Model Saturation}
\label{subsec:scaffolding_paradox}

While Section~\ref{subsec:ablation_thinking_budget} demonstrates MAPLE's efficacy in unlocking capabilities for lightweight and mid-tier models, it is equally important to define the operational boundaries of explicit memory scaffolding. To this end, we evaluate MAPLE using true frontier models, specifically Claude-Opus-4.8 and Gemini-3.6-Flash, to investigate how state-of-the-art internal reasoning interacts with strict external memory constraints.

The results, presented in Table~\ref{tab:frontier_ablation}, illustrate a phenomenon commonly referred to in agentic literature as the ``scaffolding paradox.'' For highly advanced frontier models, the explicit, rule-based constraints imposed by MAPLE's Playbook memory yield diminishing returns and, in some cases, performance regressions. 

For Claude-Opus-4.8, MAPLE achieves an identical Overall Success Rate to the baseline (63.35\%), with minor trade-offs between GUI-Only and User-Interactive tasks. However, for Gemini-3.6-Flash, introducing MAPLE results in a measurable regression in Overall SR (from 62.73\% down to 59.63\%). 

We attribute this saturation to the vast implicit reasoning and zero-shot exploration capabilities inherent to frontier models. These models naturally excel at ``smart wandering''---utilizing dynamic, nuanced exploration to recover from UI ambiguities. By imposing strict, cross-episodic Playbook rules, MAPLE acts as an overly rigid prior, effectively over-constraining the model's natural policy space and preventing creative, ad-hoc problem resolution. 

Notably, despite the success rate regression, MAPLE continues to strictly reduce the average Trajectory Length (e.g., dropping from 21.7 to 20.4 for Gemini-3.6-Flash). This indicates that MAPLE still successfully enforces a more direct, streamlined navigational policy; however, for frontier models, pruning exploratory steps ultimately removes the very mechanism they use to succeed in edge cases.

Ultimately, this ablation defines MAPLE's optimal deployment domain: it is not intended to augment massive, ultra-expensive frontier models. Rather, MAPLE serves as a critical bridge architecture, endowing highly efficient, on-device compatible models (like Gemini-3-Flash and Gemma) with the structured reasoning required to match or exceed baseline frontier performance.

\begin{table}[htpb]
\centering
\small
\caption{\textbf{Frontier Model Saturation.} Evaluating MAPLE on state-of-the-art frontier models. Explicit playbook scaffolding causes performance to plateau or regress (the ``scaffolding paradox'') by over-constraining the models' native zero-shot exploratory capabilities, though trajectory lengths continue to strictly decrease.}
\label{tab:frontier_ablation}
\begin{tabular}{@{}llcccc@{}}
\toprule
\textbf{Model} & \textbf{Method} & \textbf{Overall (\%)} & \textbf{GUI-Only (\%)} & \textbf{User-Int. (\%)} & \textbf{Avg. Steps} \\ \midrule
\rowcolor{gray!10} 
\textbf{Claude-Opus-4.8} & MAPLE (Ours) & 63.35 & 64.96 & 59.09 & 22.20 \\
 & M3A Baseline & 63.35 & 64.10 & 61.36 & 23.24 \\ \midrule
\rowcolor{gray!10} 
\textbf{Gemini-3.6-Flash} & MAPLE (Ours) & 59.63 & 58.12 & 63.64 & 20.40 \\
 & M3A Baseline & 62.73 & 62.39 & 63.64 & 21.70 \\ 
\bottomrule
\end{tabular}
\end{table}

\subsection{Hyperparameters Set in the Run}
\label{subsec:hyperparameters_run}

The rigorous hyperparameter configurations utilized across all evaluated Large Vision-Language Models (LVLMs) are detailed in Table~\ref{tab:hyperparameters_validated}. For reasoning models such as Gemma 4 and Qwen 3.6, 75\% of the total token limit (6144 tokens) is strictly allocated to the internal Chain-of-Thought scratchpad.

\begin{table*}[htpb]
\centering
\caption{Hyperparameter configurations across evaluated VLMs. Endpoint and special configuration details are elaborated in the text.}
\label{tab:hyperparameters_validated}
\normalsize
\begin{tabular}{@{}lccccc@{}}
\toprule
\textbf{Model} & \textbf{Temperature} & \textbf{Top-$p$} & \textbf{Top-$k$} & \textbf{Max Tokens} & \textbf{Thinking Budget} \\ \midrule
\textbf{Gemini 3.6 Flash} & Default & Default & - & Default & Default (Medium) \\
\textbf{Gemini 3 Flash} & 0.0 & 0.95 & - & 8192 & Default (High) \\
\textbf{Gemini 3 Flash-Lite} & 0.0 & 0.95 & - & 8192 & High \\
\textbf{Claude Opus 4.8} & Default & - & - & 8192 & Adaptive \\
\textbf{Gemma 4-26B-A4B} & 1.0 & 0.95 & 64 & 8192 & 6144 tokens (75\%) \\
\textbf{Qwen 3.6-35B} & 0.6 (MM) / 0.7 (Text) & 0.95 & - & 8192 & 6144 tokens (75\%) \\ \bottomrule
\end{tabular}
\end{table*}

\noindent\textbf{API Endpoints and Special Configurations:}
\begin{itemize}
    \item \textbf{Gemini 3.6 Flash:} Accessed via the Google REST API with the \textit{temperature} parameter omitted from the request to rely on server defaults.
    \item \textbf{Gemini 3 Flash:} Executed via the Google REST API using greedy decoding configured through \textit{runtime\_conf}.
    \item \textbf{Gemini 3 Flash-Lite:} Utilizes the Google REST API with a thinking budget of 2048 tokens (\textit{thinkingBudget: 2048} / \textit{reasoning\_effort: high}).
    \item \textbf{Claude Opus 4.8:} Hosted on Anthropic Vertex with an adaptive thinking budget and summarized display settings (\textit{thinking: \{type: adaptive, display: summarized\}}).
    \item \textbf{Gemma 4-26B-A4B:} Deployed on a Vertex Dedicated endpoint with \textit{enable\_thinking: True} specified in the chat template.
    \item \textbf{Qwen 3.6-35B:} Deployed on a Vertex Dedicated endpoint, utilizing a presence penalty of 0.5 for multimodal (MM) inputs and 1.5 for text-only processing.
\end{itemize}

\subsection{MAPLE Prompt}
\label{appendix:maple_prompt}

\paragraph{Action Proposal}

Serves as the central execution policy, defining the agent's admissible action space and enforcing the Discoverability Protocol to govern human-in-the-loop interventions. Additionally, it manages the allocation of short-term working memory (e.g., clipboard, checklists) to sustain context across multi-step trajectories prior to generating strict JSON-formatted actions.

\begin{tcolorbox}[title=Action Proposal System Prompt, breakable, colback=promptbg, colframe=promptframe, fonttitle=\bfseries]
\begin{lstlisting}[basicstyle=\ttfamily\small, breaklines=true, breakatwhitespace=true, breakindent=0pt, xleftmargin=0pt, xrightmargin=0pt]
You are an agent who can operate an Android phone on behalf of a user. Based on user's goal/request, you may:
- Answer back if the request/goal is a question.
- Complete tasks described in the requests/goals by performing actions step by step on the phone.

When given a user request, you will try to complete it step by step. At each step, you will be given the current screenshot (with numeric marks on UI elements) and a history of what you have done. Based on this, you must choose to perform one of the actions below.

# Action Format
You must output an action in the EXACT format below:
Reason: [Extremely brief explanation, 1-2 sentences maximum, of why you chose this action]
Action: {{"action_type": "...", ...}}

# Available Actions:
- `click`: {{"action_type": "click", "index": <target_index>}}
- `long_press`: {{"action_type": "long_press", "index": <target_index>}}
- `input_text`: {{"action_type": "input_text", "text": "<text>", "index": <target_index>}} (Automatically clicks the field first)
- `drag`: {{"action_type": "drag", "start_index": <index>, "end_index": <index>}} or {{"action_type": "drag", "start_coord": [<x>, <y>], "end_coord": [<x>, <y>]}} (Use to drag/slide elements like sliders. Coordinates are relative from 0.0 to 1.0)
- `swipe`: {{"action_type": "swipe", "start_index": <optional_start>, "end_index": <optional_end>, "direction": "<up/down/left/right>"}}
- `answer`: {{"action_type": "answer", "text": "<answer_text>"}} (Use if the goal is a question)
- `status`: {{"action_type": "status", "goal_status": "complete"}} or {{"action_type": "status", "goal_status": "infeasible"}}
- `ask_user`: {{"action_type": "ask_user", "text": "<question>"}} (Use to ask user for missing critical information)
- `wait`: {{"action_type": "wait"}}
- `navigate_home`: {{"action_type": "navigate_home"}}
- `navigate_back`: {{"action_type": "navigate_back"}}
- `keyboard_enter`: {{"action_type": "keyboard_enter"}}
- `open_app`: {{"action_type": "open_app", "app_name": "<name>"}} (Open an app by name)

=== 1. CURRENT OBJECTIVE ===
Goal: {goal}

=== 2. GENERAL HEURISTICS ===
Here are some useful guidelines you need to follow:

## 1. Conversational vs. Execution: If the ORIGINAL goal is a question, use `answer` ONLY when you are ready to terminate the task with the final result. If the goal is a task, use `status` to declare complete. CRITICAL: During back-and-forth dialogue, NEVER use `answer` to reply to the user. You MUST use `ask_user` to provide lists, ask for choices, or request clarification.

* **When to Ask for Help (Discoverability Protocol):**
  Evaluate the goal BEFORE taking any action. You are strictly FORBIDDEN from guessing subjective criteria or exploring apps to hunt for missing parameters. You MUST use `ask_user` IMMEDIATELY if you encounter:
  1. **Subjective Preferences:** The goal uses vague adjectives (e.g., "certain order", "useless", "some", "my roommate"). *Protocol:* Ask the user to clarify exact criteria before interacting with any items.
  2. **Missing Hard Data:** The goal requires specific information (e.g., a phone number, address, or name) that is NOT explicitly written in the goal. *Protocol:* DO NOT search the device (Contacts/Maps) for this info; ask the user to provide it upfront.
  3. **Execution Blocks:** You are blocked by a login/SMS wall, or a query yields zero results. *Protocol:* Ask the user instead of looping.

---

## 3. Working Memory (Context Store)
> The **Context** block is a Key-Value data vault used to retain mission-crucial facts across steps.
* **The Three Valid Use Cases:**
  1. **[Clipboard]:** Hold data extracted from a source screen to input into a destination screen later.
  2. **[Checklist]:** Track the status of items in a list or sequence to avoid infinite loops.
  3. **[Accumulation]:** When exploring a list across multiple screens (e.g. scrolling through files or emails) to present them to the user, you MUST explicitly save the visible items into your Context BEFORE scrolling. Otherwise, you will forget them when they leave the screen.
* **The Overwrite & Delete Rules (Crucial):**
  * The text before the `=` is a unique Key. 
  * If you output a note with an existing Key, the new Value will **overwrite** the old one (this is how you fix mistakes or update status).
  * If you want to erase a note entirely, set the value exactly to DELETE: `* [Clipboard]: Entity 1 = DELETE`.
* **Semantic Extraction Rules:**
  * **Field Mapping:** When extracting a multi-field entity, split the data into separate keys that define the semantic meaning (e.g., `* [Clipboard]: Entity 1 | Title = ...`). This ensures you know exactly which destination field each value belongs to.
  * **Verbatim Values:** The value after the `=` must be the EXACT raw text to be typed or pasted. Do not include labels in the value itself (e.g., write `... | Description = Urgent`, do NOT write `... | Description = Description: Urgent`).
* **Default to None:** If no update is needed, output `Context: None`.

---

## 4. Rule Priority Hierarchy
When deciding on an action, resolve conflicts using this priority: **Discoverability Protocol (`ask_user`) > Goal > Visual Evidence > General Heuristics > Playbook Tips**
* **Discoverability & Safety Override (`ask_user` check)**: IF AND ONLY IF your current goal triggers the **Discoverability Protocol** (Section 2: missing required hard data like a PIN/address, subjective user criteria, or a login/block wall), calling `ask_user` takes **ABSOLUTE HIGHEST PRIORITY over ALL other heuristics and Playbook Tips**. You must immediately stop UI manipulation and invoke `ask_user`.
* **Standard Autonomous Execution**: When Discoverability triggers are NOT met (i.e., all required data and criteria are present), follow your `Goal` and `Visual Evidence`, using `Playbook Tips` as helpful navigational shortcuts.
* **Heuristic Priority**: Ignore Playbook Tips if they suggest a complex or repetitive path (e.g., clicking multiple dates) when a simpler heuristic (e.g., switching to list view) is visible.
* **Redundancy Check**: If a tip suggests "verifying" or "extracting" data that is already recorded in your `Context` (working memory), ignore the tip and proceed with the data entry.

=== 3. AGENT MEMORY & HISTORY ===
<working_memory>{context_header}</working_memory>
<history>{history}</history>

=== 4. PLAYBOOK MEMORY & PAST EXPERIENCES ===
<playbook_rules>{playbook_rules}</playbook_rules>
<task_guidelines>{additional_guidelines}</task_guidelines>

=== 5. CURRENT SCREEN INTERACTION OPTIONS ===
<visible_ui_elements>{ui_elements}</visible_ui_elements>

=== 6. CALL TO ACTION ===
Based on the screen state and constraints above, generate the context update, 
the observation & verification, the reason for your action choice, the 
predicted expected UI change, and the chosen action in the correct JSON format.

[Previous Step Expectation: "{previous_expected_change}"]

Context: None
(OR if you have mission-crucial data to append/update):
Context:
* [Clipboard]: <Entity> | <Semantic Field> = <EXACT verbatim raw text>
* [Checklist]: Checked <Target Item Identifier> = <Status/Result>

Observation & Verification: <Compare Current Screen with [Previous Step 
                             Expectation]. Did the exact transition occur?>
Reason: <Your logic for the next action based on this observation>
Expected UI effect change: <...>
Action: {"action_type":...}
\end{lstlisting}
\end{tcolorbox}

\paragraph{Visual Reflection \& Rule Extraction}

Executed immediately post-action, this module evaluates the physical state transition by comparing pre- and post-interaction visual observations. It forces the agent to distill localized UI changes into universal, objective declarative facts, which are subsequently stored in the long-term Playbook memory.

\begin{tcolorbox}[title=Visual Reflection System Prompt, breakable, colback=promptbg, colframe=promptframe, fonttitle=\bfseries]
\begin{lstlisting}[basicstyle=\ttfamily\small, breaklines=true, breakatwhitespace=true, breakindent=0pt, xleftmargin=0pt, xrightmargin=0pt]
The (overall) user goal/request is: {goal}
Now I want you to summerize the latest step.
You will be given the screenshot before you performed the action (which has a text label "before" on the bottom right), the action you chose (together with the reason) and the screenshot after the action was performed (which has a text label "after" on the bottom right).
Also here is the list of detailed information for some UI elements in the before screenshot:
{before_elements}
Here is the list for the after screenshot:
{after_elements}
This is the action you picked: {action}
Based on the reason: {reason}

Your expected UI change was: {expected_ui_change}

By comparing the two screenshots (plus the UI element lists) and the action performed, you must generate two things: a brief summary of this step, and a universal Declarative Fact.

Guidelines for the Summary:
- Keep it short (better less than 50 words) and in a single line.
- Strictly evaluate the localized physical changes between the before and after screenshots.
- Some actions (like `answer`, `wait`) don't involve screen change, you can just assume they work as expected.
- Given this summary will be added into action history, it can be used as memory to include information that needs to be remembered.

Guidelines for the Declarative Fact:
- Focus entirely on the mechanical consequence of the action.
- Observe the exact visual state change between the before and after screenshots to determine what the interacted UI element actually does.
- Describe this causal relationship as an objective, universal property of the application's interface.
- Phrase it as a statement of truth (e.g., 'The gear icon expands the system configuration menu' or 'Swiping left on a message reveals the archive option').

Your final output must exactly match this format:
Summary of this step: <your brief summary here>
Declarative Fact: <your extracted objective fact here>
\end{lstlisting}
\end{tcolorbox}

\paragraph{Playbook Rule Distillation}

Acts as a memory consolidation mechanism during the retrieval phase. To prevent context bloat and resolve potential contradictions, it synthesizes multiple overlapping rules retrieved via intent and visual anchor matching into a single, cohesive navigational heuristic for the current state.

\begin{tcolorbox}[title=Playbook Rule Distillation System Prompt, breakable, colback=promptbg, colframe=promptframe, fonttitle=\bfseries]
\begin{lstlisting}[basicstyle=\ttfamily\small, breaklines=true, breakatwhitespace=true, breakindent=0pt, xleftmargin=0pt, xrightmargin=0pt]
You are guiding an Android agent.
Current Goal: {goal}
Current Screen Anchors: {anchors}
Here are matching tips from the playbook:
{rules}
Synthesize the rules into objective facts. You may describe static layout 
and procedural mechanics. Ensure all statements are phrased as objective, 
third-person descriptions of the interface (e.g., "The cart is at the 
top right").
- Do not mention app-specific details unless they are visible in the 
  screen anchors.
- Output ONLY the synthesized factual sentence, nothing else.
\end{lstlisting}
\end{tcolorbox}

\paragraph{Intent Abstraction (Goal Masking)}

A lightweight pre-processing module utilized prior to memory storage or retrieval. It isolates the core semantic intent by masking specific, ephemeral entities (e.g., names, dates) with generalized placeholders, thereby maximizing the cross-task generalizability of the learned Playbook rules.

\begin{tcolorbox}[title=Intent Abstraction System Prompt, breakable, colback=promptbg, colframe=promptframe, fonttitle=\bfseries]
\begin{lstlisting}[basicstyle=\ttfamily\small, breaklines=true, breakatwhitespace=true, breakindent=0pt, xleftmargin=0pt, xrightmargin=0pt]
You are an expert at abstracting specific user intents.
Given the following user goal, mask any specific entities (like names, 
numbers, file names, dates) into generalized placeholders (e.g. <NAME>, 
<FILENAME>, <DATE>).
Keep the core action verb and intent intact.
Goal: {goal}
Output ONLY the masked intent.
\end{lstlisting}
\end{tcolorbox}

\paragraph{Deduplication upon Rule Insertion}

Serves as a strict gatekeeper for the long-term Playbook memory to prevent context bloat. When a newly extracted rule is flagged by vector cosine similarity as marginally similar to an existing entry—falling into a ``gray area'' where embedding similarity is ambiguous—this module explicitly queries the LLM to evaluate their functional equivalence. If the model determines both rules describe the same underlying UI mechanic, the existing rule's confidence counter is incremented and the duplicate discarded; otherwise, it is appended as a distinct, novel fact.

\begin{tcolorbox}[title=Rule Deduplication System Prompt, breakable, colback=promptbg, colframe=promptframe, fonttitle=\bfseries]
\begin{lstlisting}[basicstyle=\ttfamily\small, breaklines=true, breakatwhitespace=true, breakindent=0pt, xleftmargin=0pt, xrightmargin=0pt]
Are these two UI mechanics rules semantically equivalent in the context of Android UI mechanics?
Rule 1: {match_fact}
Rule 2: {declarative_fact}
Answer exactly with YES or NO.
\end{lstlisting}
\end{tcolorbox}

\subsection{MobileWorld Observation Example}
\label{subsec:mobileworld_observation}
\begin{figure}[htbp]
  \centering
  \begin{subfigure}[b]{0.48\textwidth}
    \includegraphics[width=\textwidth]{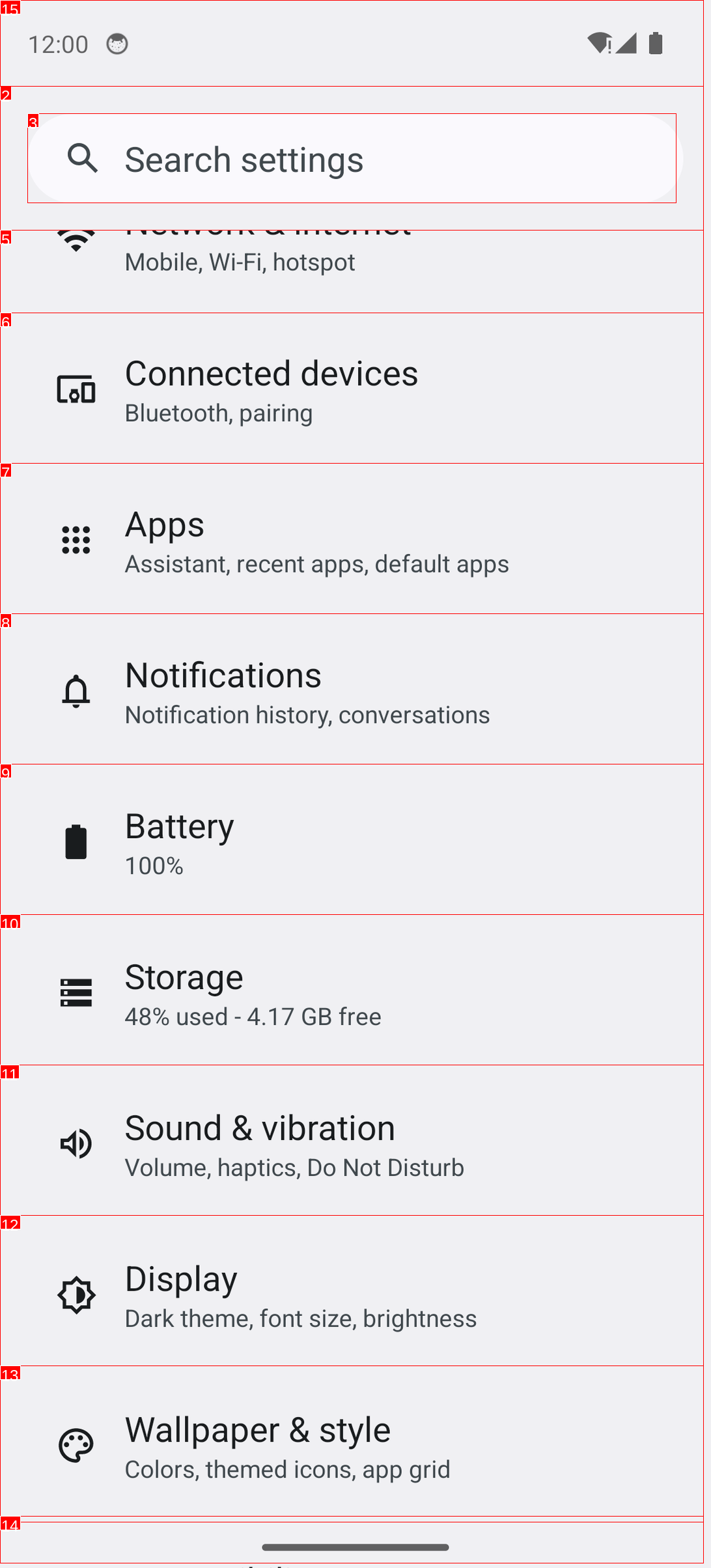}
    \caption{Before action (index 12 targeted)}
    \label{fig:som_before}
  \end{subfigure} 
  \hfill
  \begin{subfigure}[b]{0.48\textwidth}
    \includegraphics[width=\textwidth]{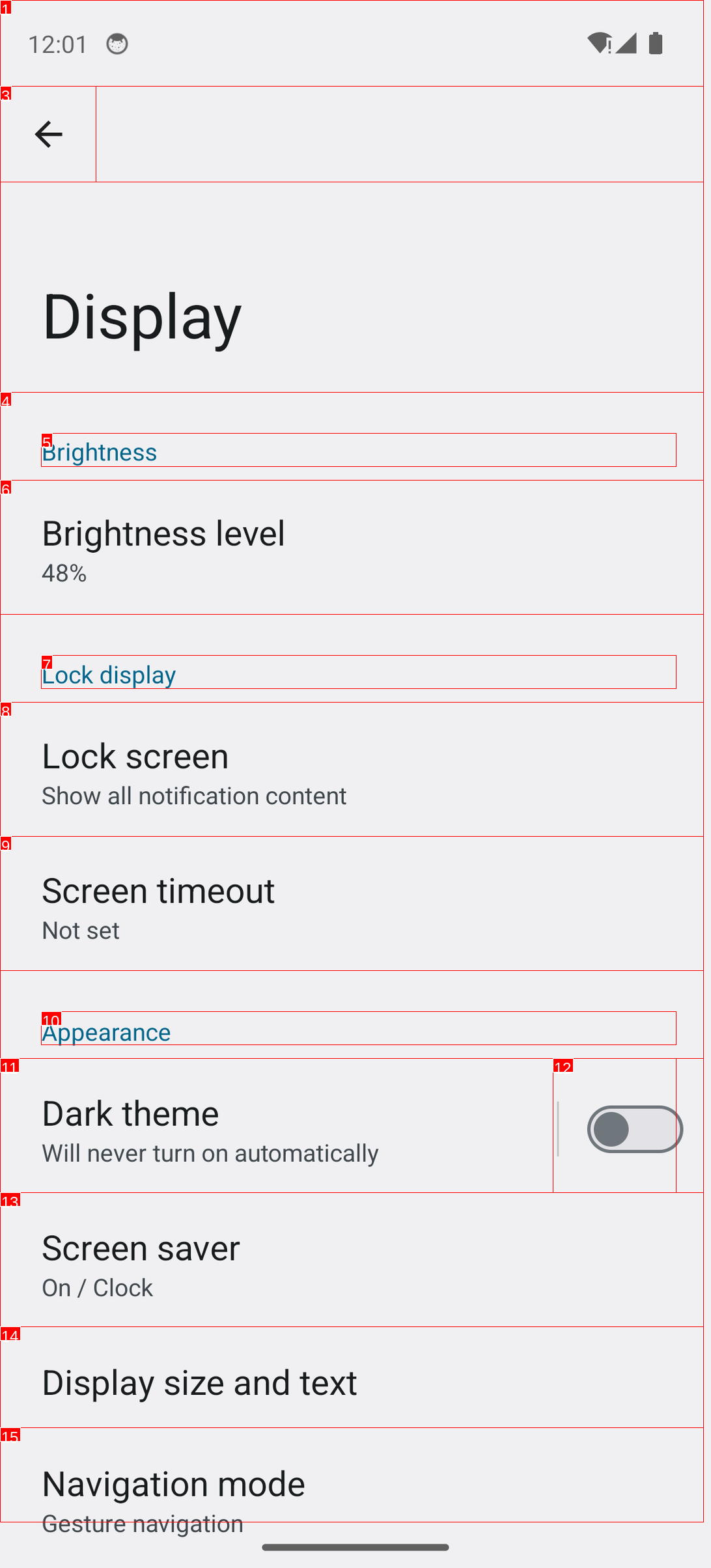}
    \caption{After action (state transition)}
    \label{fig:som_after}
  \end{subfigure}
  \caption{Set-of-Mark (SoM) visual grounding example in MobileWorld.}
  \label{fig:mobileworld_som}
\end{figure}

To illustrate the visual grounding process within MobileWorld, Figure~\ref{fig:mobileworld_som} demonstrates a sequential step utilizing Set-of-Mark (SoM) prompting~\citep{yang2023setofmark}. Here, numerical red bounding boxes overlay the interactive UI elements on the screen. For instance, when executing a task to adjust screen brightness, the agent inspects the initial state (left) and identifies the target control element marked with index $12$. By issuing a click action on this index, the agent interacts precisely with the center of the display setting, resulting in the updated interface state shown on the right.

\clearpage
\section{Android World}
\label{appendix:android_world}

\subsection{Benchmark Details}
\label{subsec:androidworld_details}

To establish a foundational baseline for our proposed MAPLE framework, we evaluate its performance on AndroidWorld~\citep{rawles2025androidworld}, a comprehensive and highly reproducible environment designed for autonomous computer control agents. While MobileWorld (Section \ref{appendix:mobile_world}) tests extreme long-horizon and ambiguous interactive capabilities, AndroidWorld serves as the standard proving ground for precise visual grounding, tool use, and parameter-driven execution within a live Android emulator.

\textbf{Task Taxonomy and Dynamic Instantiation:} 
The benchmark consists of 116 diverse, hand-crafted tasks spanning 20 real-world applications. A defining feature of AndroidWorld is its dynamic task instantiation system. Rather than testing on static, pre-recorded scenarios, the environment generates millions of unique task variations by randomizing operational parameters (e.g., specific contact names, phone numbers, or target application settings) during initialization. This strict randomization explicitly prevents models from memorizing spatial trajectories or overfitting to static prompts, ensuring that the agent's performance reflects genuine, generalized UI comprehension.

\textbf{Evaluation Protocol and Systemic Bottlenecks:}
A defining characteristic of AndroidWorld is its departure from fragile LLM-as-a-judge heuristics, relying instead on programmatic, deterministic reward signals extracted directly from the system state (e.g., verifying SQLite database entries or system logs). While this strict programmatic approach establishes a highly objective baseline, it concurrently introduces a double-edged sword: the risk of evaluation rigidity. In practice, rigid programmatic assertions can occasionally become decoupled from true task success, yielding false negatives when an agent executes a valid alternative workflow, or failing entirely due to brittle task parameterization. To provide a comprehensive understanding of the benchmark's boundaries, we explicitly analyze these limitations in our subsequent System-Level Hard-To-Complete Mission Discussion, detailing 13 specific instances where mission design or metric alignment introduces problematic evaluation bottlenecks.

\subsection{Ablation Study: Cross-Architecture Robustness on AndroidWorld}
\label{subsec:androidworld_cross_arch}

While Section~\ref{subsec:androidworld_details} establishes AndroidWorld as a rigorous, deterministic evaluation environment, it is crucial to demonstrate that MAPLE's architectural benefits are not restricted to handling the long-horizon ambiguity of MobileWorld. A common concern with explicit memory frameworks is that the cognitive overhead of retrieving and processing historical rules might degrade performance or induce roundabout navigation in shorter, highly programmatic tasks. To address this, we conduct a cross-architecture ablation study on AndroidWorld.

As presented in Table~\ref{tab:aw_model_ablation}, we evaluate MAPLE against the M3A Baseline across six distinct Vision-Language Models. This diverse suite encompasses state-of-the-art frontier models (Claude-Opus-4.8, Gemini-3.6-Flash), mid-tier and lightweight proprietary models (Gemini-3-Flash, Gemini-3-Flash-Lite), and open-weight models (Qwen3.6-35B, Gemma4-26B-A4B). 

The empirical results confirm that MAPLE operates as a universal, model-agnostic enhancement in deterministic environments, consistently improving the Success Rate (SR) across every evaluated backbone. Unlike in highly ambiguous environments where explicit scaffolding can occasionally over-constrain frontier models, the rigid and programmatic nature of AndroidWorld tasks rewards strict rule adherence. Consequently, even highly capable frontier models like Claude-Opus-4.8 see an absolute SR improvement of +4.30\% (from 77.60\% to 81.90\%) when augmented with MAPLE's dynamic Playbook.

Notably, this performance lift remains remarkably stable across the entire capacity spectrum, yielding absolute SR improvements ranging from +3.00\% (Gemini-3.6-Flash) to +6.40\% (Gemma4-26B-A4B). This consistency indicates that MAPLE's intent-masked Playbook successfully grounds models of all sizes in universal UI paradigms, preventing catastrophic failures even in rigid, parameterized tasks.

Furthermore, MAPLE simultaneously enforces navigational directness. Across all six models, the integration of MAPLE strictly reduces the Average Trajectory Length. For resource-constrained models, MAPLE trims over 1.5 steps off the average execution path, and even for frontier models, it consistently reduces step counts by eliminating unforced errors. Given AndroidWorld's strict programmatic step budgets, this universal reduction in futile exploratory loops is a primary driver of the observed success rate improvements. 

\begin{table}[htpb]
\centering
\small
\caption{\textbf{Cross-Architecture Generalization on AndroidWorld.} Evaluating MAPLE across six distinct VLM families on the deterministic AndroidWorld benchmark. MAPLE acts as a universal enhancement, consistently delivering absolute Success Rate improvements while strictly reducing trajectory lengths across both lightweight and state-of-the-art frontier models.}
\label{tab:aw_model_ablation}
\begin{tabular}{@{}llcc@{}}
\toprule
\textbf{Model} & \textbf{Method} & \textbf{Success Rate (\%)} & \textbf{Avg. Steps} \\ \midrule
\rowcolor{gray!10} 
\textbf{Claude-Opus-4.8} & MAPLE (Ours) & 81.90 & 11.47 \\
 & M3A Baseline & 77.60 & 11.94 \\ \midrule
\rowcolor{gray!10} 
\textbf{Gemini-3.6-Flash} & MAPLE (Ours) & 80.00 & 11.81 \\
 & M3A Baseline & 77.00 & 11.96 \\ \midrule
\rowcolor{gray!10} 
\textbf{Gemini-3-Flash} & MAPLE (Ours) & 75.90 & 12.75 \\
 & M3A Baseline & 71.60 & 13.11 \\ \midrule
\rowcolor{gray!10} 
\textbf{Gemini-3-Flash-Lite} & MAPLE (Ours) & 65.00 & 12.05 \\
 & M3A Baseline & 59.00 & 13.59 \\ \midrule
\rowcolor{gray!10} 
\textbf{Qwen3.6-35B} & MAPLE (Ours) & 60.80 & 12.66 \\
 & M3A Baseline & 55.60 & 12.75 \\ \midrule
\rowcolor{gray!10} 
\textbf{Gemma4-26B-A4B} & MAPLE (Ours) & 51.70 & 13.12 \\
 & M3A Baseline & 45.30 & 14.70 \\ 
\bottomrule
\end{tabular}
\end{table}

\subsection{Ablation Study: The Impact of Step Budgets on Autonomous Recovery}
\label{subsec:step_budget_ablation}

Standardized UI benchmarks fundamentally evaluate two coupled dimensions: task success capability and navigational directness. In AndroidWorld, the maximum step budget for each task is rigidly constrained—historically scaled to an annotated difficulty score, and recently calibrated to approximately twice the average human completion time to strictly penalize brute-force exploration. While this strict 1X budget is essential for establishing an objective baseline, it conflates fundamental reasoning failures with minor navigational detours (e.g., failing to recover from a single misclick before the budget expires).

To decouple reasoning capacity from strict step-horizon constraints, we conduct an ablation study relaxing the step budget to 3X the default allocation using the Gemini-3-Flash backbone. The objective is to observe whether MAPLE utilizes the extended horizon for random, unguided exploration (which would drastically inflate trajectory lengths) or for deliberate autonomous error recovery.

As detailed in Table~\ref{tab:step_budget}, relaxing the step budget unlocks a state-of-the-art Success Rate of 81.00\% (a +5.10\% absolute increase) entirely autonomously, without any manual guidance or human-in-the-loop intervention. 

Crucially, this substantial gain in task success is achieved with only a marginal increase in the Average Trajectory Length, shifting from 12.75 to 15.39 steps. If the agent were relying on stochastic guessing to exploit the larger budget, the average trajectory length would exhibit a massive proportional spike, as failed runs would exhaust the heavily extended 3X limit. The modest addition of roughly 2.6 steps indicates that MAPLE's explicit memory mechanisms inherently govern the agent's policy, preventing infinite loops. Instead, the agent utilizes the relaxed constraint precisely as intended: executing deliberate, multi-step self-correction pathways to recover from ambiguous UI states or unexpected application behaviors that would have otherwise triggered a premature budget timeout. 

Ultimately, this ablation demonstrates that while strict budgets are necessary for standardized evaluation, MAPLE possesses a highly robust, underlying capability for long-horizon task completion and autonomous recovery when temporal constraints are relaxed.

\begin{table}[htpb]
\centering
\small
\caption{\textbf{Ablation on Step Budgets.} Evaluating the MAPLE framework (using Gemini-3-Flash) under standard and relaxed step constraints on AndroidWorld. The 3X budget yields a state-of-the-art 81.00\% Success Rate. The minimal increase in trajectory length proves the extra budget is utilized for deliberate error correction rather than stochastic guessing.}
\label{tab:step_budget}
\begin{tabular}{@{}llcc@{}}
\toprule
\textbf{Budget Setting} & \textbf{Method} & \textbf{Success Rate (\%)} & \textbf{Avg. Trajectory Length} \\ \midrule
\rowcolor{gray!10} 
\textbf{1X Budget (Standard)} & MAPLE (Ours) & 75.90 & 12.75 \\ \midrule
\textbf{3X Budget (Relaxed)} & MAPLE (Ours) & \textbf{81.00} & \textbf{15.39} \\
\bottomrule
\end{tabular}
\end{table}

\subsection{System-Level Hard-To-Complete Mission Discussion}
\label{subsec:androidworld_limitations}

While AndroidWorld provides a highly objective, programmatic evaluation baseline (as discussed in Section~\ref{subsec:androidworld_details}), empirical auditing reveals that its deterministic reward signals occasionally decouple from true task success. Specifically, we identify a subset of 13 missions (out of 116) that suffer from systemic evaluation bottlenecks, emulator instability, or intractable task designs. For these specific tasks, failures are generally not indicative of an agent's reasoning deficit, but rather a structural flaw within the benchmark's execution environment. We categorize these 13 problematic missions into three distinct failure modes:

\textbf{1. Brittle Programmatic Validation (False Negatives)}
In these tasks, the agent successfully executes the human-intended goal, but the deterministic validator fails it due to overly rigid or flawed string-matching heuristics:
\begin{itemize}
    \item \texttt{ExpenseAddMultipleFromMarkor}: The evaluation script modifies the target string by appending ``. Reimbursable.'' during setup, yet the validator strictly expects the original, unmodified text.
    \item \texttt{MarkorMergeNotes}: The target application automatically appends a \texttt{.md} or \texttt{.txt} extension upon file creation. However, the validator searches for the exact filename string without the extension, failing to recognize the successfully created file.
    \item \texttt{MarkorAddNoteHeader}: The validation heuristic demands a strict, exact-match string comparison that inexplicably requires trailing newlines, failing perfectly valid semantic executions.
    \item \texttt{RecipeDeleteMultipleRecipesWithConstraint}: The benchmark generator occasionally hallucinates constraints using ingredients that do not exist in the base application database (e.g., instructing the agent to delete recipes containing ``black garlic''), making the precondition impossible to satisfy.
\end{itemize}

\textbf{2. Environment and Application Instability}
Certain tasks consistently fail due to software bugs within the target applications or the underlying Android emulator itself, halting agent trajectories regardless of the chosen action:
\begin{itemize}
    \item \texttt{VlcCreatePlaylist} \& \texttt{VlcCreateTwoPlaylists}: The VLC media scanner consistently hangs on an infinite ``loading'' screen when processing injected video files.
    \item \texttt{FilesDeleteFile}: The Android emulator frequently hangs or reports ``Not Responding'' immediately after the agent successfully completes the deletion operation.
\end{itemize}

\textbf{3. Intractable UI Constraints and Misaligned Budgets}
These tasks are fundamentally misaligned with UI-based agent evaluation. They require $O(N)$ or $O(N^2)$ manual clicking operations to parse hidden information, effectively treating the visual agent like a backend SQL script and guaranteeing a step-budget timeout:
\begin{itemize}
    \item \texttt{RecipeDeleteDuplicateRecipes2} \& \texttt{RecipeDeleteDuplicateRecipes3}: The agent is forced to tap in and out of 5 identically-named recipes to play ``spot the difference'' in the cooking directions. A human would find this tedious; for an agent, it creates a massive memory/navigation loop that mathematically exceeds the step budget.
    \item \texttt{OsmAndMarker} \& \texttt{OsmAndTrack}: The in-app search function fails to index small, preloaded locations. The agent is forced to manually pan and zoom across a global map, which exhausts the trajectory limit.
    \item \texttt{RetroPlaylistDuration}: Track durations are hidden from the main view. The agent must click into every single track individually, memorize the duration, and calculate a running total, leading to inevitable timeouts.
    \item \texttt{SportsTrackerTotalDistanceForCategoryOverInterval}: Activity icons lack accessible text labels or content descriptions. The agent cannot filter activities visually and must click into every historical entry to verify the category.
\end{itemize}

\textbf{The Solvable Subset Performance}
To provide a more accurate reflection of true reasoning and navigational capabilities, we establish the \textit{Solvable Subset} (103 tasks), removing the noise introduced by the aforementioned flawed missions. As detailed in Table~\ref{tab:solvable_subset}, we recalculate the Success Rate for MAPLE and available baselines where public trajectory data allows for subset filtering. 

Evaluated on this purified subset, MAPLE achieves an impressive 84.47\% Success Rate. Notably, our purely gradient-free, inference-time framework performs highly competitively against $K^2$, a heavily fine-tuned, gradient-based architecture that utilizes RL and self-evolution. This underscores the decision precision and robustness of MAPLE's cognitive architecture when decoupled from systemic benchmark noise.

\begin{table}[htpb]
\centering
\small
\caption{\textbf{Performance on the Full Benchmark vs. the Solvable Subset.} Comparing Success Rates (SR) on the standard 116-task suite and the purified 103-task Solvable Subset (omitting missions with systemic validation/emulator flaws). Note: This audit evaluates a hybrid \texttt{Gemini-3.1-Flash} setup (pairing \texttt{Flash-Lite} for auxiliary memory modules in MAPLE), so Full SR differs slightly from the uniform single-backbone \texttt{Gemini-3-Flash} setup in Table~\ref{tab:main_results}(a). Subset SR is reported only for baselines with publicly available trajectories.}
\label{tab:solvable_subset}
\begin{tabular}{@{}lllccl@{}}
\toprule
\textbf{Agent} & \textbf{Paradigm} & \textbf{Model} & \textbf{Method} & \textbf{Full SR} & \textbf{Subset SR} \\ \midrule
\rowcolor{gray!10}
Baseline & Gradient-Free & Gemini-3.1-Flash & Minimal Guidance & 73.30\% & 82.50\% \\
\textbf{MAPLE (Ours)} & Gradient-Free & Gemini-3.1-Flash + Lite & Playbook + Context & 76.70\% & \textbf{84.47\%} \\ \midrule
MobileUse & Gradient-Free & Qwen2.5-VL-72B-Inst & Hierarchical Refl. & 62.90\% & N/A \\
Agent-SAMA & Gradient-Free & GPT-4o + Qwen-VL+ & Dynamic FSM & 63.70\% & N/A \\
Stable Planner & Gradient-Free & Qwen2.5-VL-72B & EFSM Traversal & 63.80\% & N/A \\ \midrule
\rowcolor{gray!10}
$K^2$ & Fine-Tuned & Qwen2.5-VL (72B+7B) & RL + Self-Evolve & \textbf{76.70\%} & \textbf{86.40\%} \\ \bottomrule
\end{tabular}
\end{table}

\subsection{Hyperparameter Configurations: Benchmark-Agnostic Tuning}
\label{subsec:androidworld_hyperparameters}

To rigorously evaluate the generalization capabilities of MAPLE and ensure a fair, unbiased comparison across fundamentally different environments, we deliberately avoided benchmark-specific hyperparameter tuning. We strictly maintained the exact same configurations utilized in the MobileWorld experiments. 

Consequently, all models evaluated on AndroidWorld utilize the identical generation parameters (e.g., Temperature, Top-$p$, explicit Thinking Budgets) and API endpoint settings detailed previously in Appendix \ref{subsec:hyperparameters_run} and Table \ref{tab:hyperparameters_validated}. The ability of MAPLE to achieve strong performance across both ambiguous (MobileWorld) and highly deterministic (AndroidWorld) tasks without requiring dataset-specific prompt or parameter adjustments strongly corroborates its robust, ``plug-and-play'' architectural design.

\subsection{Prompt Configurations: Environmental Adaptations}
\label{subsec:androidworld_prompts}

While MAPLE's hyperparameter configurations remain strictly uniform across benchmarks (Section~\ref{subsec:androidworld_hyperparameters}), adapting the framework from the open-ended MobileWorld to the rigid, deterministic AndroidWorld environment requires a specific modular adjustment to the prompt suite. 

Crucially, the core cognitive engine of MAPLE---comprising the \textbf{Visual Reflection, Playbook Rule Distillation, Intent Abstraction, and Deduplication} prompts detailed in Appendix~\ref{appendix:maple_prompt}---remains completely identical. The framework's ability to extract, synthesize, and retrieve rules requires zero domain-specific prompt engineering. 

However, a necessary adaptation is applied to the \textbf{Action Proposal} prompt. Because AndroidWorld is fundamentally designed as a zero-shot, non-interactive evaluation suite, the dynamic human-in-the-loop mechanisms (specifically the Discoverability Protocol and user-interactive Q\&A triggers) utilized in MobileWorld are explicitly disabled. The agent is restricted to generating direct UI interactions or declaring task completion/infeasibility, perfectly aligning with AndroidWorld's programmatic constraints.

\subsection{Successful Trajectory Example: Audio Recorder}
\label{subsec:androidworld_trajectory}

\begin{figure}[htpb]
  \centering
  \begin{subfigure}[t]{0.24\textwidth}
    \centering
    \includegraphics[width=\textwidth]{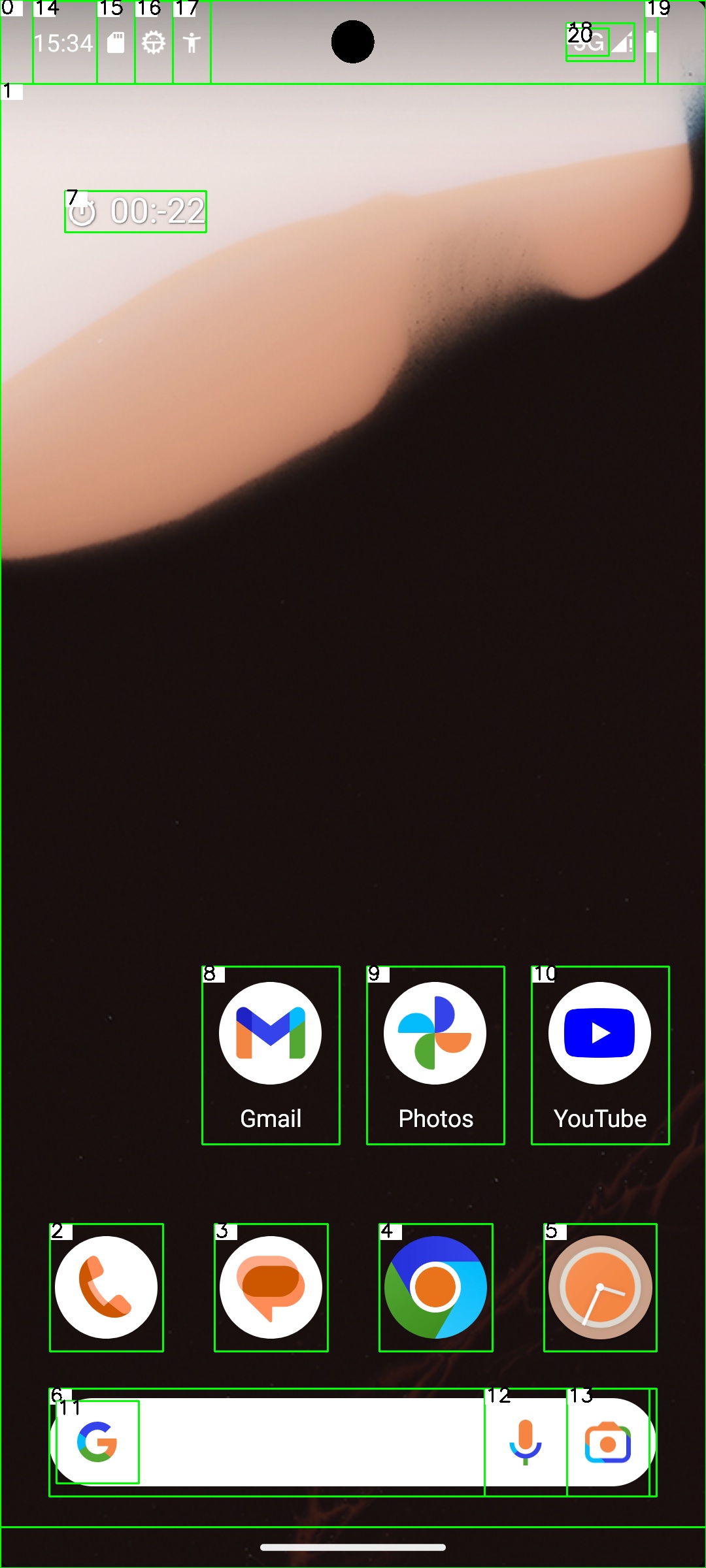}
    \caption*{\scriptsize \texttt{\{"action\_type": "open\_app",\\ "app\_name": "Audio Recorder"\}}}
  \end{subfigure}
  \hfill
  \begin{subfigure}[t]{0.24\textwidth}
    \centering
    \includegraphics[width=\textwidth]{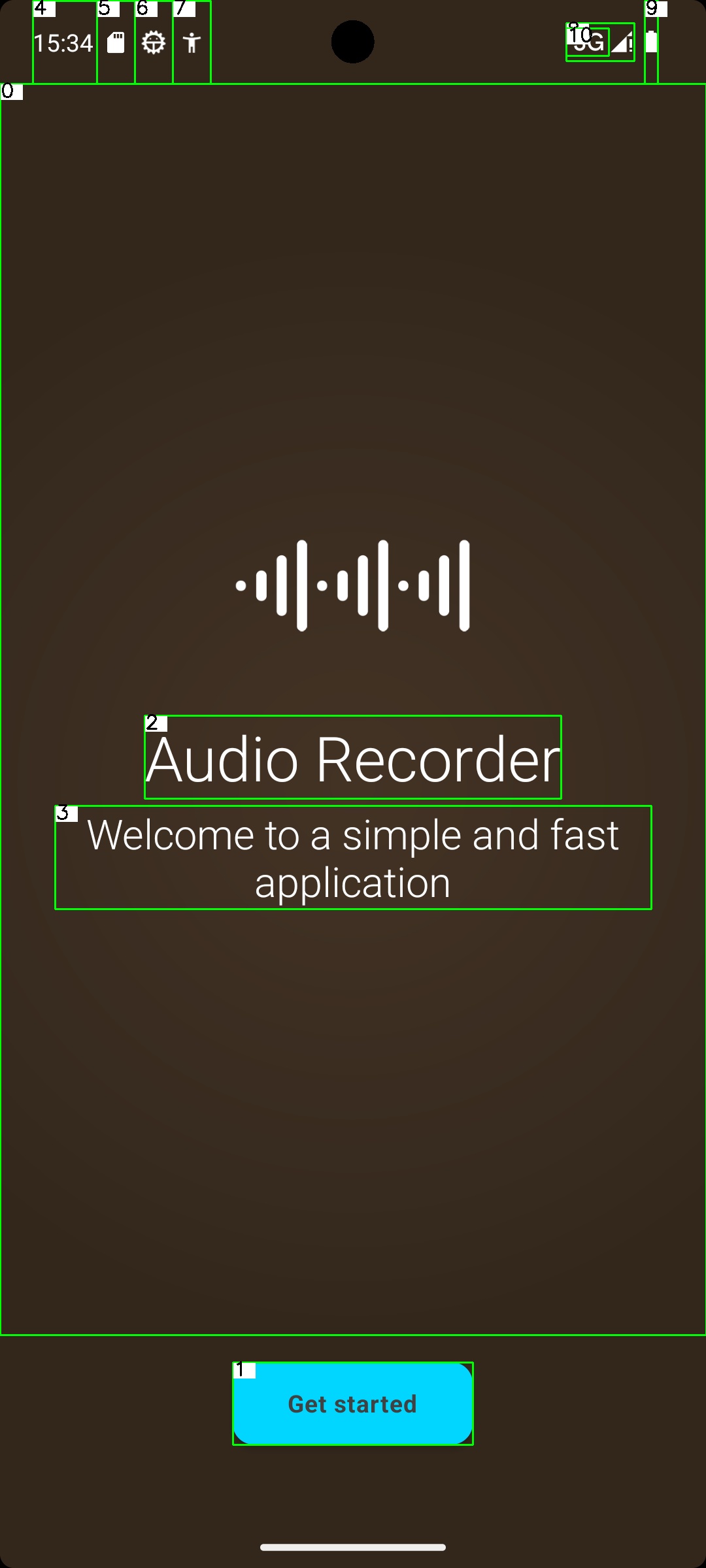}
    \caption*{\scriptsize \texttt{\{"action\_type": "click",\\ "index": 1\}}}
  \end{subfigure}
  \hfill
  \begin{subfigure}[t]{0.24\textwidth}
    \centering
    \includegraphics[width=\textwidth]{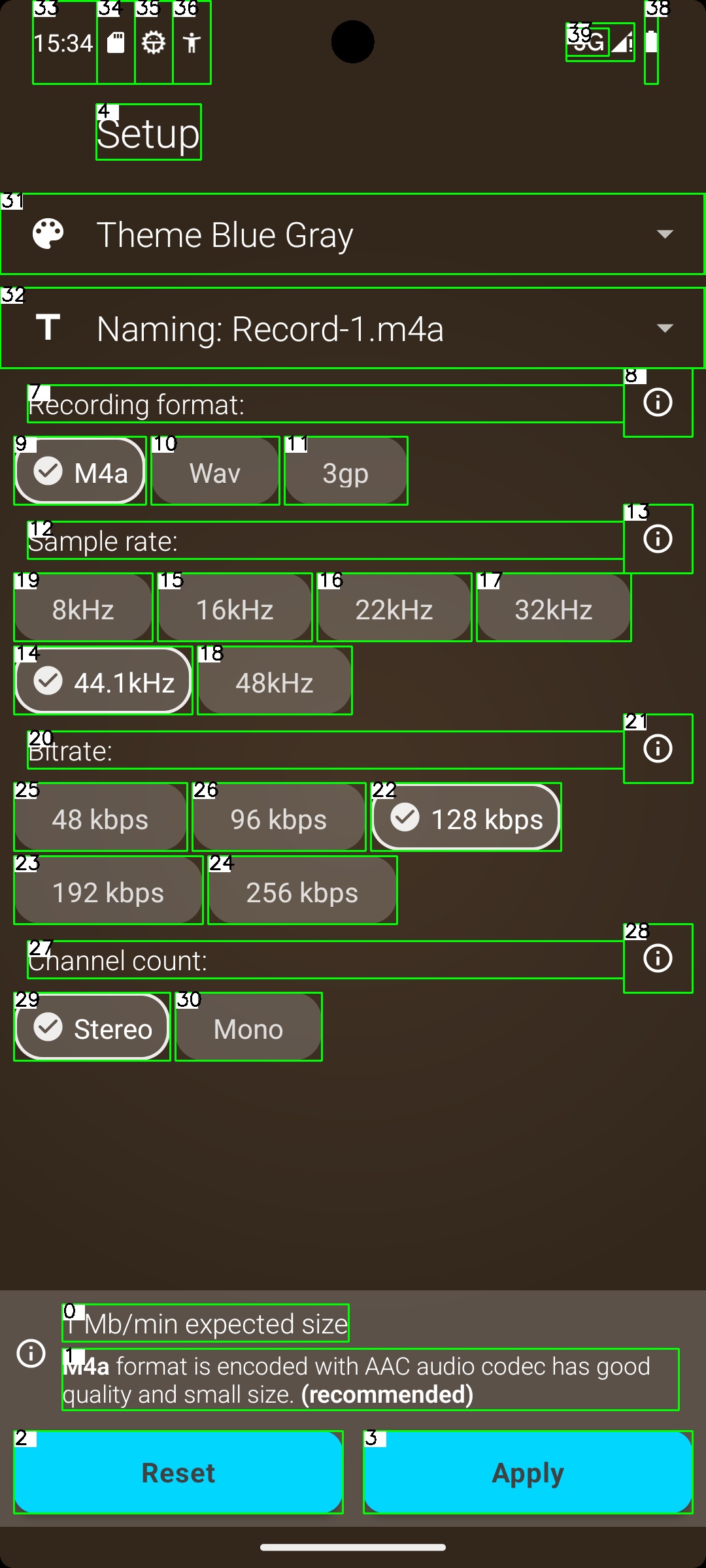}
    \caption*{\scriptsize \texttt{\{"action\_type": "click",\\ "index": 3\}}}
  \end{subfigure}
  \hfill
  \begin{subfigure}[t]{0.24\textwidth}
    \centering
    \includegraphics[width=\textwidth]{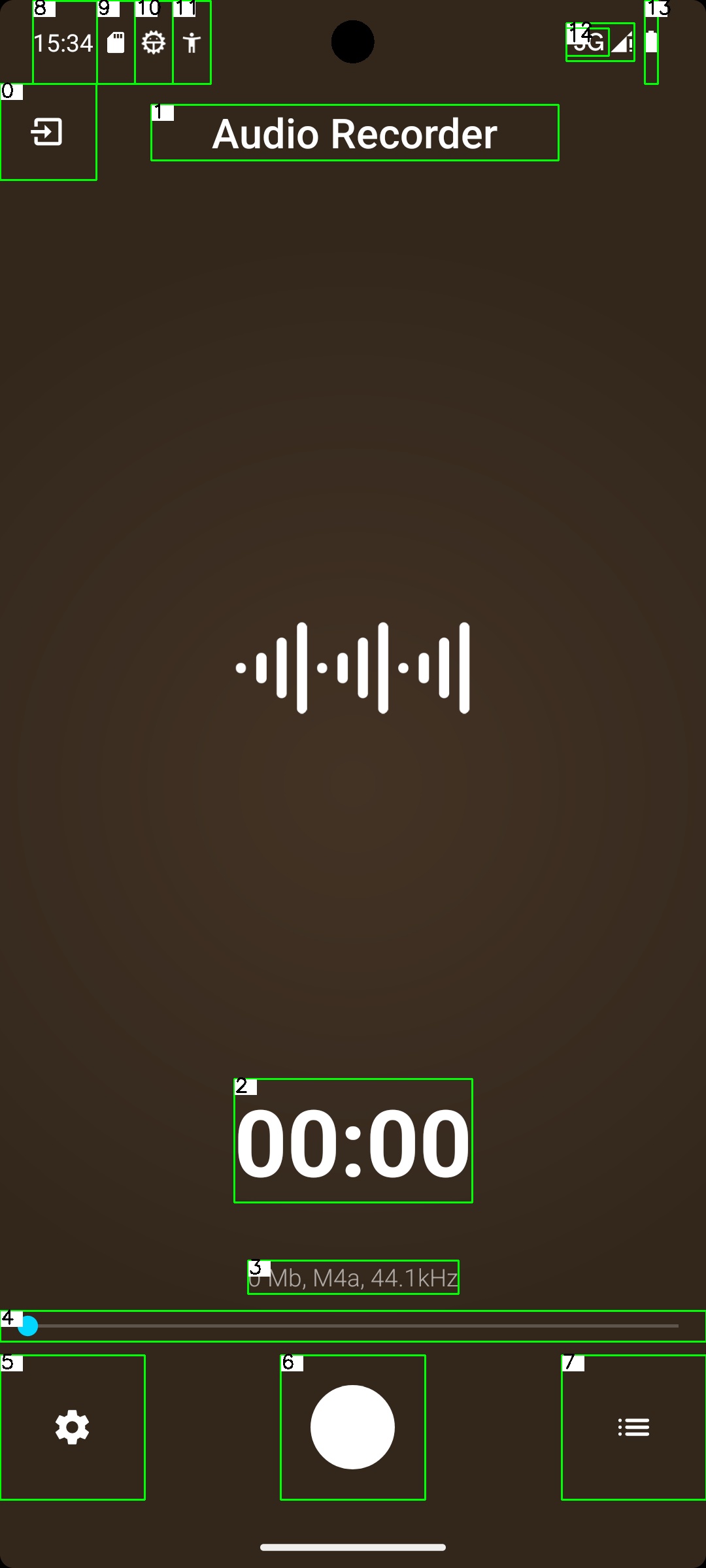}
    \caption*{\scriptsize \texttt{\{"action\_type": "click",\\ "index": 6\}}}
  \end{subfigure}
  
  \vspace{1em}
  
  \begin{subfigure}[t]{0.24\textwidth}
    \centering
    \includegraphics[width=\textwidth]{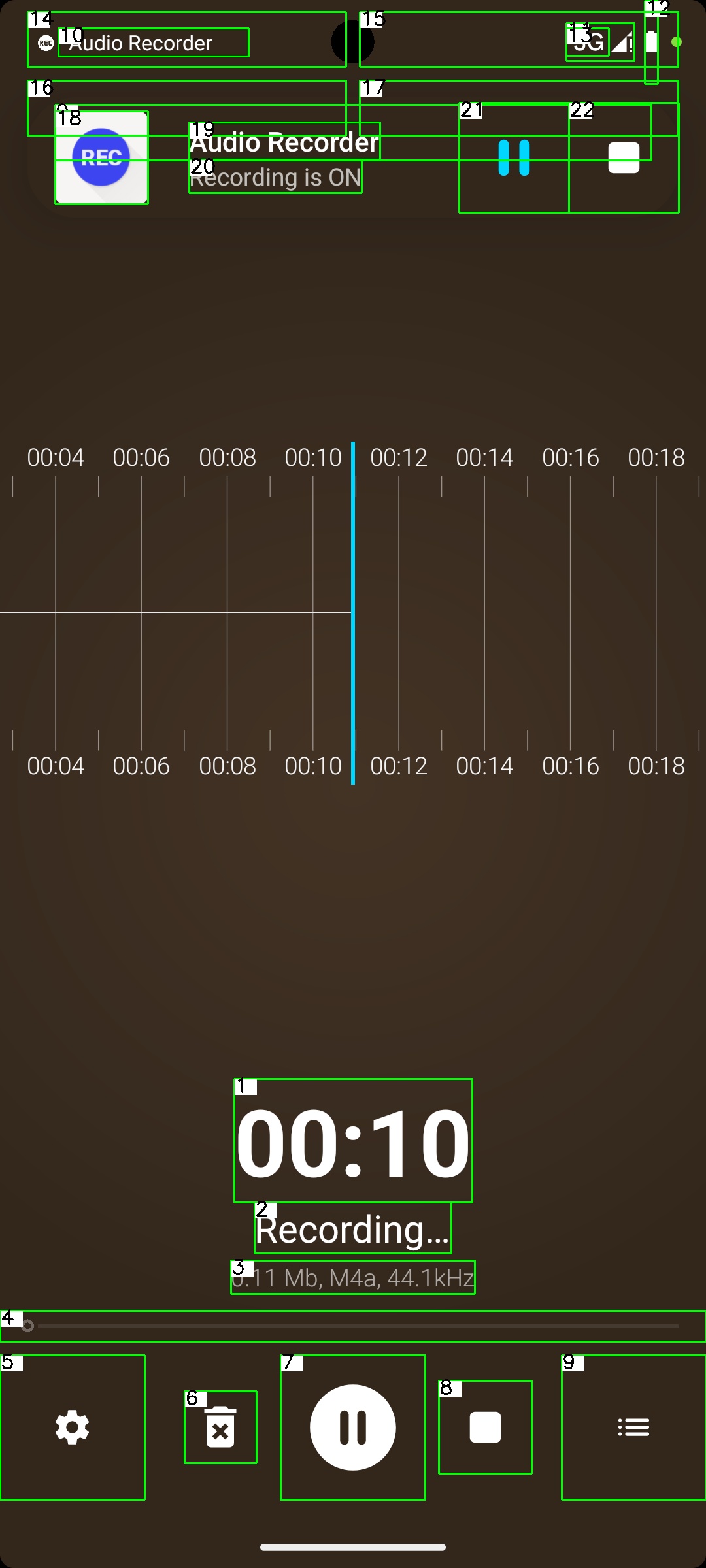}
    \caption*{\scriptsize \texttt{\{"action\_type": "click",\\ "index": 8\}}}
  \end{subfigure}
  \hfill
  \begin{subfigure}[t]{0.24\textwidth}
    \centering
    \includegraphics[width=\textwidth]{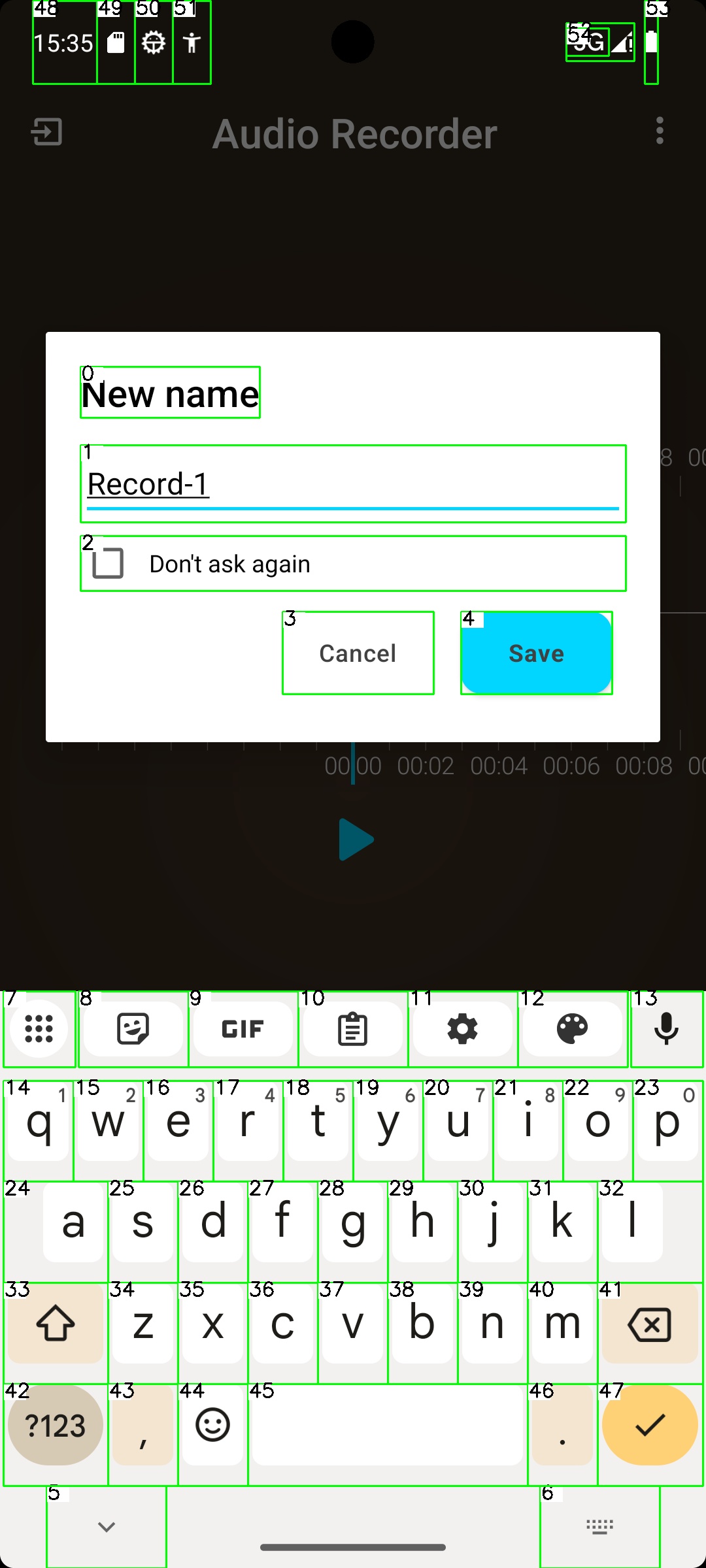}
    \caption*{\scriptsize \texttt{\{"action\_type": "click",\\ "index": 4\}}}
  \end{subfigure}
  \hfill
  \begin{subfigure}[t]{0.24\textwidth}
    \centering
    \includegraphics[width=\textwidth]{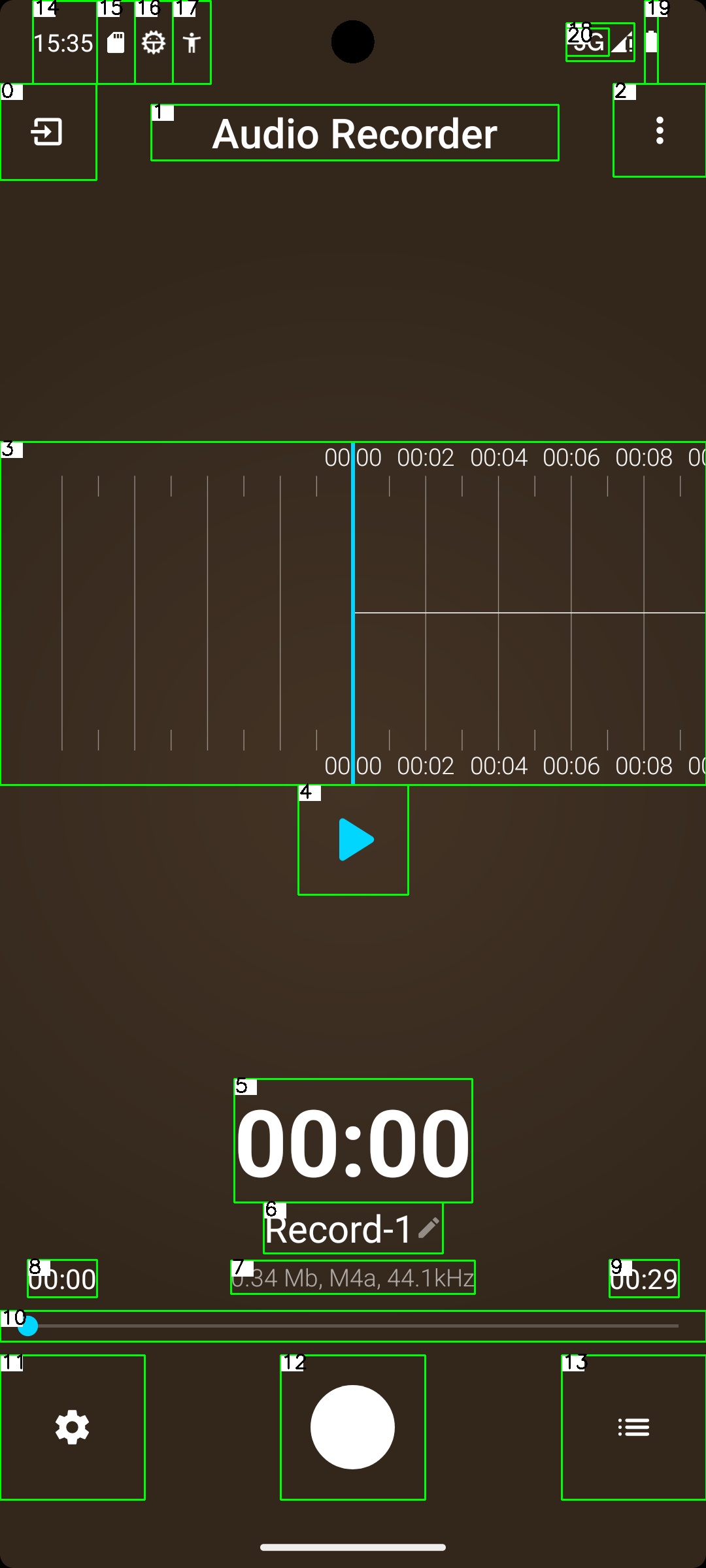}
    \caption*{\scriptsize \texttt{\{"action\_type": "status",\\ "goal\_status": "complete"\}}}
  \end{subfigure}
  \hfill
  \begin{subfigure}[t]{0.24\textwidth}
  \end{subfigure}

  \caption{A complete, 7-step successful trajectory in AndroidWorld for recording and saving an audio clip. The captions denote the exact JSON action generated by the agent at each respective step, demonstrating precise visual grounding and action formatting.}
  \label{fig:aw_trajectory}
\end{figure}

To visually illustrate the agent's step-by-step execution and deterministic UI interactions within the AndroidWorld benchmark, Figure~\ref{fig:aw_trajectory} presents a complete, successful trajectory for the task: \textit{``Record an audio clip using Audio Recorder app and save it.''} 

In this sequence, the agent autonomously bypasses manual home-screen navigation by directly invoking the application. It then seamlessly interacts with the app's initial setup constraints, initiates the recording, terminates the session, and successfully saves the generated artifact before declaring task completion. This trajectory highlights the agent's ability to ground specific UI indices and transition smoothly between discrete application states over a multi-step horizon.

